\documentclass{article}
\pdfoutput=1

\usepackage{microtype}
\usepackage{graphicx}
\usepackage{subfigure}
\usepackage{booktabs} % for professional tables
\usepackage[final,nonatbib]{neurips_2019}

\usepackage{amsmath,amsfonts,bm}
\usepackage{todonotes}
\usepackage{url}
\usepackage{algorithm,algorithmic}
\usepackage{bbm}
\usepackage{multirow}
\usepackage{wrapfig}
\usepackage{enumitem}
\usepackage{xspace}
\usepackage{eqparbox}
\usepackage{setspace}

\usepackage{amsmath,amsfonts,bm}

\def\eqref#1{equation~\ref{#1}}
\def\1{\bm{1}}

\DeclareMathAlphabet{\mathsfit}{\encodingdefault}{\sfdefault}{m}{sl}
\SetMathAlphabet{\mathsfit}{bold}{\encodingdefault}{\sfdefault}{bx}{n}

\newcommand{\softmax}{\mathrm{softmax}}

\newcommand{\diag}{\mathrm{diag}}

\newcommand{\tb}[1]{\textbf{#1}}
\usepackage{hyperref}
\newcommand{\ourmodel}{Attention Attractor Networks}

\newcommand{\ourproblem}{Incremental Few-Shot Learning}
\newcommand{\ourproblemsmall}{incremental few-shot learning}
\newcommand{\ua}{\uparrow}
\newcommand{\da}{\downarrow}
\newcommand{\D}{$\Delta \da$}
\newcommand{\Da}{$\Delta_a \da$}
\newcommand{\Db}{$\Delta_b \da$}

\makeatletter
\DeclareRobustCommand\onedot{\futurelet\@let@token\@onedot}
\def\@onedot{\ifx\@let@token.\else.\null\fi\xspace}

\def\eg{\emph{e.g}\onedot} 
\def\ie{\emph{i.e}\onedot}

\def\arxiv{1}
\makeatother
\begin{document}
\title{{\ourproblem} with {\ourmodel}}
\author{Mengye Ren$^{1,2,3}$, Renjie Liao$^{1,2,3}$, Ethan Fetaya$^{1,2}$, Richard S. Zemel$^{1,2}$\\
${}^1$University of Toronto, ${}^2$Vector Institute, ${}^3$Uber ATG\\
\texttt{\{mren, rjliao, ethanf, zemel\}@cs.toronto.edu}}
\maketitle
\vspace{-0.2in}
% !TEX root = ../main.tex
\begin{abstract}
Machine learning classifiers are often trained to recognize a set of pre-defined classes. However,
in many applications, it is often desirable to have the flexibility of learning additional concepts,
with limited data and without re-training on the full training set. This paper addresses this
problem, {\it \ourproblemsmall}, where a regular classification network has already been trained to
recognize a set of base classes, and several extra novel classes are being considered, each with
only a few labeled examples. After learning the novel classes, the model is then evaluated on the
overall classification performance on both base and novel classes. To this end, we propose a
meta-learning model, the Attention Attractor Network, which regularizes the learning of novel
classes. In each episode, we train a set of new weights to recognize novel classes until they
converge, and we show that the technique of recurrent back-propagation can back-propagate through
the optimization process and facilitate the learning of these parameters. We demonstrate that the
learned attractor network can help recognize novel classes while remembering old classes without the
need to review the original training set, outperforming various baselines.
\end{abstract}
\vspace{-0.1in}
% !TEX root = ../main.tex
\section{Introduction}
The availability of large scale datasets with detailed annotation, such as
ImageNet~\cite{imagenet}, played a significant role in the recent success of deep learning. The
need for such a large dataset is however a limitation, since its collection requires intensive human
labor. This is also strikingly different from human learning, where new concepts can be learned from
very few examples. One line of work that attempts to bridge this gap is few-shot
learning~\cite{koch2015siamese,matching,proto}, where a model learns to output a classifier given
only a few labeled examples of the unseen classes. While this is a promising line of work, its
practical usability is a concern, because few-shot models only focus on learning novel classes,
ignoring the fact that many common classes are readily available in large datasets.

An approach that aims to enjoy the best of both worlds, the ability to learn from large datasets for
common classes with the flexibility of few-shot learning for others, is {\it \ourproblemsmall}
\cite{lwof}. This combines incremental learning where we want to add new classes without
catastrophic forgetting \cite{mccloskey1989catastrophic}, with few-shot learning when the new
classes, unlike the base classes, only have a small amount of examples. One use case to illustrate
the problem is a visual aid system. Most objects of interest are common to all users, e.g., cars,
pedestrian signals; however, users would also like to augment the system with additional
personalized items or important landmarks in their area. Such a system needs to be able to learn new
classes from few examples, without harming the performance on the original classes and typically
without access to the dataset used to train the original classes.

In this work we present a novel method for incremental few-shot learning where during meta-learning
we optimize a regularizer that reduces catastrophic forgetting from the incremental few-shot
learning. Our proposed regularizer is inspired by attractor networks \cite{localist} and can be
thought of as a memory of the base classes, adapted to the new classes. We also show how this
regularizer can be optimized, using recurrent back-propagation \cite{rbp,rbp2,rbp3} to
back-propagate through the few-shot optimization stage. Finally, we show empirically that our
proposed method can produce state-of-the-art results in incremental few-shot learning on {\it
mini}-ImageNet \cite{matching} and {\it tiered}-ImageNet \cite{fewshotssl} tasks.
% !TEX root = ../main.tex
\begin{figure}[t]
\centering
\includegraphics[width=0.7\textwidth,trim={1cm 0 0 0.7cm},clip]{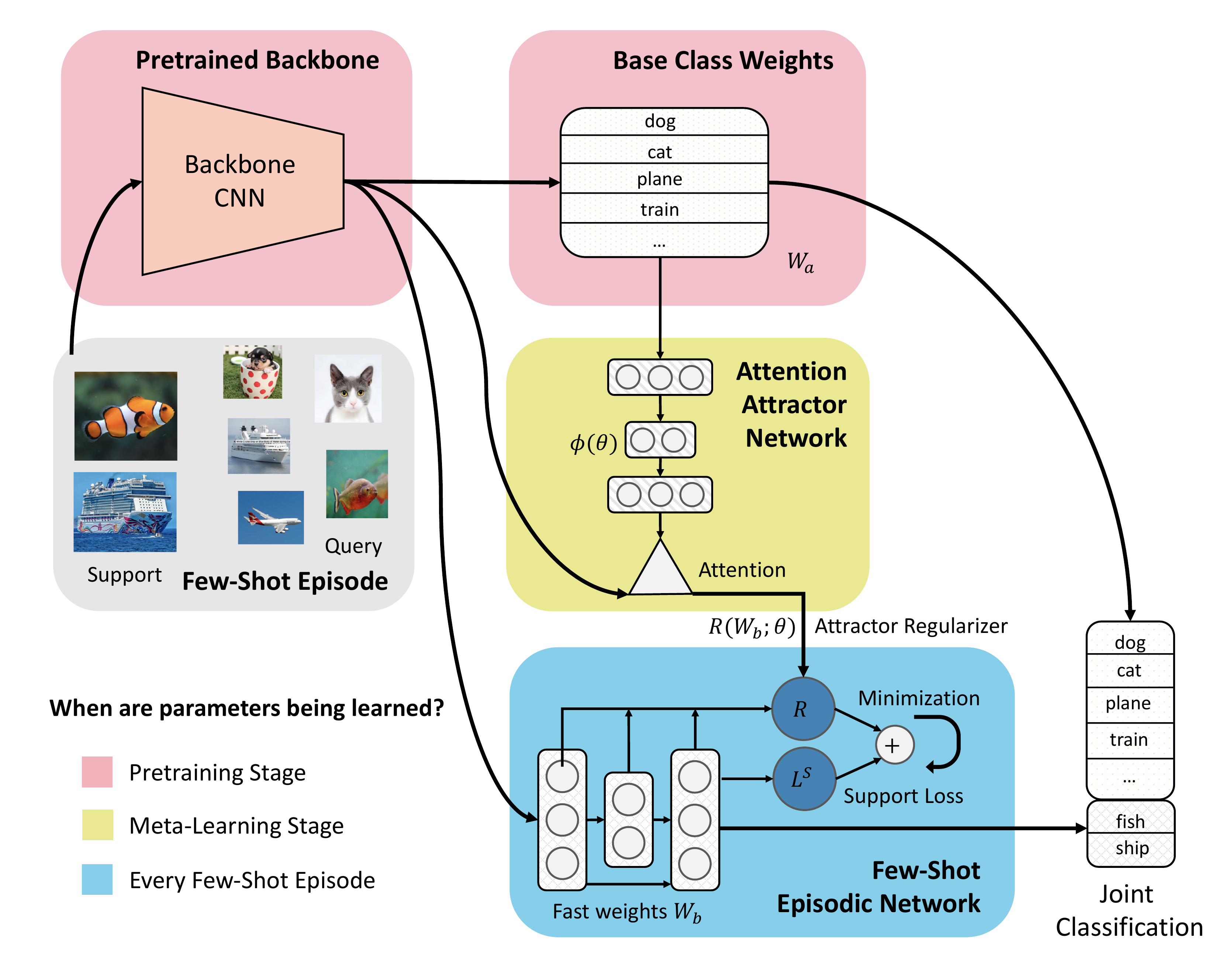}
\caption{Our proposed attention attractor network for incremental few-shot learning.
During pretraining we learn the base class weights $W_a$ and the feature extractor CNN backbone. In
the meta-learning stage, a few-shot episode is presented. The support set only contains novel
classes, whereas the query set contains both base and novel classes. We learn an episodic classifier
network through an iterative solver, to minimize cross entropy plus an additional regularization
term predicted by the attention attractor network by attending to the base classes. The attention
attractor network is meta-learned to minimize the expected query loss. During testing an episodic
classifier is learned in the same way.}
\label{fig:model}
\end{figure}
% !TEX root = ../main.tex
\section{Related Work}
Recently, there has been a surge in interest in few-shot learning
\cite{koch2015siamese,matching,proto,lake2011oneshot}, where a model for novel classes is learned
with only a few labeled examples. One family of approaches for few-shot learning, including Deep
Siamese Networks~\cite{koch2015siamese}, Matching Networks~\cite{matching} and Prototypical
Networks~\cite{proto}, follows the line of metric learning. In particular, these approaches use deep
neural networks to learn a function that maps the input space to the embedding space where examples
belonging to the same category are close and those belonging to different categories are far apart.
Recently, \cite{garcia2017few} proposes a graph neural networks based method which captures the
information propagation from the labeled support set to the query set. \cite{fewshotssl} extends
Prototypical Networks to leverage unlabeled examples while doing few-shot learning. Despite their
simplicity, these methods are very effective and often competitive with the state-of-the-art.

Another class of approaches aims to learn models which can adapt to the episodic tasks. In
particular, \cite{metalstm} treats the long short-term memory (LSTM) as a meta learner such that it
can learn to predict the parameter update of a base learner, e.g., a convolutional neural network
(CNN). MAML~\cite{maml} instead learns the hyperparameters or the initial parameters of the base
learner by back-propagating through the gradient descent steps. \cite{santoro2016one} uses a
read/write augmented memory, and \cite{mishra2017meta} combines soft attention with temporal
convolutions which enables retrieval of information from past episodes.

Methods described above belong to the general class of meta-learning models. First proposed in
~\cite{Schmidhuber1987evolutionary,naik1992meta,Thrun1998}, meta-learning is a machine learning
paradigm where the meta-learner tries to improve the base learner using the learning experiences
from multiple tasks. Meta-learning methods typically learn the update policy yet lack an overall
learning objective in the few-shot episodes. Furthermore, they could potentially suffer from
short-horizon bias \cite{shorthorizon}, if at test time the model is trained for longer steps. To
address this problem, \cite{diffsolver} proposes to use fast convergent models like logistic
regression (LR), which can be back-propagated via a closed form update rule. Compared to
\cite{diffsolver}, our proposed method using recurrent back-propagation \cite{rbp,rbp2,rbp3} is more
general as it does not require a closed-form update, and the inner loop solver can employ any
existing continuous optimizers.

Our work is also related to incremental learning, a setting where information is arriving
continuously while prior knowledge needs to be transferred. A key challenge is \textit{catastrophic
forgetting}~\cite{mccloskey1989catastrophic,mcclelland1995there}, i.e., the model forgets the
learned knowledge. Various memory-based models have since been proposed, which store training
examples explicitly \cite{icarl,mbpa,castro2018end,varcontinual}, regularize the parameter updates
\cite{kirkpatrick2017overcoming}, or learn a generative model \cite{fearnet}. However, in these
studies, incremental learning typically starts from scratch, and usually performs worse than a
regular model that is trained with all available classes together since it needs to learned a good
representation while dealing with catastrophic forgetting.

Incremental few-shot learning is also known as low-shot learning. To leverage a good representation,
\cite{hariharan2017lowshot,wang2018lowshot,lwof} starts off with a pre-trained network on a set of
base classes, and tries to augment the classifier with a batch of new classes that has not been seen
during training. \cite{hariharan2017lowshot} proposes the squared gradient magnitude loss, which
makes the learned classifier from the low-shot examples have a smaller gradient value when learning
on all examples. \cite{wang2018lowshot} propose the prototypical matching networks, a combination of
prototypical network and matching network. The paper also adds hallucination, which generates new
examples. \cite{lwof} proposes an attention based model which generates weights for novel
categories. They also promote the use of cosine similarity between feature representations and
weight vectors to classify images.

In contrast, during each few-shot episode, we directly learn a classifier network that is randomly
initialized and solved till convergence, unlike \cite{lwof} which directly output the prediction.
Since the model cannot see base class data within the support set of each few-shot learning episode,
it is challenging to learn a classifier that jointly classifies both base and novel categories.
Towards this end, we propose to add a learned regularizer, which is predicted by a meta-network, the
``attention attractor network''. The network is learned by differentiating through few-shot learning
optimization iterations. We found that using an iterative solver with the learned regularizer
significantly improves the classifier model on the task of incremental few-shot learning.
% !TEX root = ../main.tex
\section{Model}
In this section, we first define the setup of incremental few-shot learning, and then we introduce
our new model, the Attention Attractor Network, which attends to the set of base classes according
to the few-shot training data by using the attractor regularizing term. Figure~\ref{fig:model}
illustrates the high-level model diagram of our method.
\subsection{\ourproblem}
The outline of our meta-learning approach to incremental few-shot learning is: (1) We learn a fixed
feature representation and a classifier on a set of base classes; (2) In each training and testing
episode we train a novel-class classifier with our meta-learned regularizer; (3) We optimize our
meta-learned regularizer on combined novel and base classes classification, adapting it to perform
well in conjunction with the base classifier. Details of these stages follow.

\paragraph{Pretraining Stage:} We learn a base model for the regular supervised classification task
on dataset $\{(x_{a,i},y_{a,i})\}_{i=1}^{N_a}$ where $x_{a,i}$ is the $i$-th example from dataset
$\mathcal{D}_a$ and its labeled class $y_{a,i}\in\{1,2,...,K\}$. The purpose of this stage is to
learn both a good base classifier and a good representation. The parameters of the base classifier
are learned in this stage and will be fixed after pretraining. We denote the parameters of the top
fully connected layer of the base classifier $W_a \in \mathbb{R}^{D\times K}$ where $D$ is the
dimension of our learned representation.
\paragraph{Incremental Few-Shot Episodes:} A few-shot dataset $\mathcal{D}_b$ is presented, from
which we can sample few-shot learning episodes $\mathcal{E}$. Note that this can be the same data
source as the pretraining dataset $\mathcal{D}_a$, but sampled episodically. For each $N$-shot
$K'$-way episode, there are $K'$ novel classes disjoint from the base classes. Each novel class has
$N$ and $M$ images from the support set $S_b$ and the query set $Q_b$ respectively. Therefore, we
have $\mathcal{E} = (S_b, Q_b), S_b = (x_{b,i}^S, y_{b,i}^S)_{i=1}^{N \times K'}, Q_b = (x_{b,i}^Q,
y_{b,i}^Q)_{i=1}^{M
\times K'}$ where $y_{b,i} \in \{K+1,...,K+K'\}$. $S_b$ and $Q_b$ can be regarded as this episodes
training and validation sets. Each episode we learn a classifier on the support set $S_b$ whose
learnable parameters $W_b$ are called the \textit{fast weights} as they are only used during this
episode. To evaluate the performance on a joint prediction of both base and novel classes, i.e., a
$(K+K')$-way classification, a mini-batch $Q_a=\{(x_{a,i}, y_{a,i})\}_{i=1}^{M \times K}$ sampled
from $\mathcal{D}_a$ is also added to $Q_b$ to form $Q_{a+b} = Q_a \cup Q_b$. This means that the
learning algorithm, which only has access to samples from the novel classes $S_b$, is evaluated on
the \emph{joint} query set $Q_{a+b}$.
\paragraph{Meta-Learning Stage:} In meta-training, we iteratively sample few-shot episodes
$\mathcal{E}$ and try to learn the meta-parameters in order to minimize the joint prediction loss on
$Q_{a+b}$. In particular, we design a regularizer $R(\cdot, \theta)$ such that the \textit{fast
weights} are learned via minimizing the loss $\ell(W_b,S_b)+R(W_b, \theta)$ where $\ell(W_b,S_b)$ is
typically cross-entropy loss for few-shot classification. The meta-learner tries to learn
meta-parameters $\theta$ such that the optimal \textit{fast weights} $W_b^*$ w.r.t. the above loss
function performs well on $Q_{a+b}$. In our model, meta-parameters $\theta$ are encapsulated in our
attention attractor network, which produces regularizers for the fast weights in the few-shot
learning objective.

\paragraph{Joint Prediction on Base and Novel Classes:} We now introduce the details of our joint
prediction framework performed in each few-shot episode. First, we construct an episodic classifier,
\eg, a logistic regression (LR) model or a multi-layer perceptron (MLP), which takes the learned
image features as inputs and classifies them according to the few-shot classes.

During training on the support set $S_b$, we learn the \textit{fast weights} $W_{b}$ via minimizing
the following regularized cross-entropy objective, which we call the {\it episodic objective}:
\begin{equation}
\label{eq:general_form}
L^{S}(W_{b}, \theta) = - \frac{1}{NK'}\sum_{i=1}^{N K'} \sum_{c=K+1}^{K+K'}
y_{b,i,c}^S \log \hat{y}_{b,i,c}^S  + R(W_b,\theta).
\end{equation}
This is a general formulation and the specific functional form of the regularization term
$R(W_b,\theta)$ will be specified later. The predicted output $\hat{y}_{b,i}^S$ is obtained via,
$\hat{y}_{b,i}^S = \softmax(\left[W_{a}^\top x_{b,i}, h(x_{b,i}; W_b^\ast) \right])$, where
$h(x_{b,i})$ is our classification network and $W_{b}$ is the fast weights in the network. In the
case of LR, $h$ is a linear model: $h(x_{b,i}; W_{b}) = W_b^\top x_{b,i}$. $h$ can also be an MLP
for more expressive power.

During testing on the query set $Q_{a+b}$, in order to predict both base and novel classes, we
directly augment the softmax with the fixed base class weights $W_a$, $\hat{y}_{i}^Q =
\softmax(\left[W_{a}^\top x_{i}, h(x_{i}; W_b^\ast) \right])$, where ${W}_{b}^{\ast}$ are the
optimal parameters that minimize the regularized classification objective in
Eq.~(\ref{eq:general_form}).

\subsection{\ourmodel}
Directly learning the few-shot episode, e.g., by setting $R(W_b,\theta)$ to be zero or simple
weight decay, can cause catastrophic forgetting on the base classes. This is because $W_b$ which is
trained to maximize the correct novel class probability can dominate the base classes in the joint
prediction. In this section, we introduce the Attention Attractor Network to address this problem.
The key feature of our attractor network is the regularization term $R(W_b, \theta)$:
% \vskip -5mm
\begin{equation}
R(W_b, \theta) = 
\sum_{k^{\prime} = 1}^{K^{\prime}} 
(W_{b,k^{\prime}} - u_{k^{\prime}})^\top \diag(\exp(\gamma)) (W_{b,k^{\prime}} - u_{k^{\prime}}),
\end{equation}
where $u_{k^{\prime}}$ is the so-called \textit{attractor} and $W_{b,k^{\prime}}$ is the
$k^{\prime}$-th column of $W_b$. This sum of squared Mahalanobis distances from the attractors adds
a bias to the learning signal arriving solely from novel classes. Note that for a classifier such as
an MLP, one can extend this regularization term in a layer-wise manner. Specifically, one can have
separate attractors per layer, and the number of attractors equals the number of output dimension
of that layer.

To ensure that the model performs well on base classes, the attractors $u_{k^{\prime}}$ must contain
some information about examples from base classes. Since we can not directly access these base
examples, we propose to use the \textit{slow weights} to encode such information. Specifically, each
base class has a learned attractor vector $U_k$ stored in the memory matrix $U=[U_1,...,U_K]$. It is
computed as, $U_k = f_{\phi}(W_{a, k})$, where $f$ is a MLP of which the learnable parameters are
$\phi$. For each novel class $k^{\prime}$ its classifier is regularized towards its attractor
$u_{k^{\prime}}$ which is a weighted sum of $U_k$ vectors. Intuitively the weighting is an attention
mechanism where each novel class attends to the base classes according to the level of interference,
i.e. how prediction of new class $k'$ causes the forgetting of base class $k$.

For each class in the support set, we compute the cosine similarity between the average
representation of the class and  base weights $W_a$ then normalize using a softmax function
\begin{align}
a_{k^{\prime}, k} = 
\frac{\exp \left(\tau A(\frac{1}{N}\sum_{j} h_j \mathbbm{1}[y_{b,j} = 
k^{\prime}], W_{a, k}) \right)}
{\sum_{k} \exp \left(\tau A(\frac{1}{N}\sum_{j} h_j \mathbbm{1}[y_{b,j} = 
k^{\prime}], W_{a, k}) \right)},
\end{align}
where $A$ is the cosine similarity function, $h_j$ are the representations of the inputs in the
support set $S_b$ and $\tau$ is a learnable temperature scalar. $a_{k^{\prime},k}$ encodes a
normalized pairwise attention matrix between the novel classes and the base classes. The attention
vector is then used to compute a linear weighted sum of entries in the memory matrix $U$,
$u_{k^{\prime}} = \sum_k a_{k^{\prime}, k} U_k + U_0$, where $U_0$ is an embedding vector and serves
as a bias for the attractor.

% !TEX root = ../main.tex
\begin{wrapfigure}{R}{0.55\textwidth}
\vspace{-0.25in}
\begin{minipage}[t]{0.55\textwidth}
\begin{algorithm}[H]
\begin{small}
\caption{Meta Learning for {\ourproblem}}
\label{alg:energy}
\begin{algorithmic}[1]
\REQUIRE $\theta_0$, $\mathcal{D}_a$, $\mathcal{D}_b$, $h$
\ENSURE $\theta$
\STATE $\theta \gets \theta_0$;
\FOR{$t=1$ ... $T$}
\STATE $\{(x_b^S, y_b^S)\}, \{(x_b^Q,y_b^Q)\} \gets \text{GetEpisode}(\mathcal{D}_b)$;
\STATE $\{x_{a+b}^Q, y_{a+b}^Q\} \gets \text{GetMiniBatch}(\mathcal{D}_a) \cup \{(x_b^Q, y_b^Q)\}$;
\STATE $ $
\REPEAT
    \STATE $L^S \gets \frac{1}{NK'} \sum_i y_{b,i}^S \log \hat{y}_{b,i}^S + R(W_b; \theta)$;
    \STATE $W_b \gets \text{OptimizerStep}(W_b, \nabla_{W_b} L^S)$;
\UNTIL{$W_b$ \text{converges}}

\STATE $\hat{y}_{a+b,j}^Q \gets \softmax([W_a^\top x_{a+b, j}^Q, h( x_{a+b, j}^Q;W_b) ])$;
\STATE $L^Q \gets \frac{1}{2NK'} \sum_j y_{a+b,j}^Q \log \hat{y}_{a+b,j}^Q$;
\STATE ~\\
\COMMENT{Backprop through the above optimization via RBP}\\
\COMMENT{A dummy gradient descent step}
\STATE $W_b' \gets W_b - \alpha \nabla_{W_b} L^S$;% \ \ \ \COMMENT{A dummy gradient descent step}
\STATE $J \gets \frac{\partial W_b'}{\partial W_b}$; $v \gets \frac{\partial L^Q}{\partial W_b}$; $g \gets v$;

\REPEAT
\STATE $v \gets J^\top v - \epsilon v$; $g \gets g + v$;
\UNTIL{$g$ \text{converges}}
\STATE ~\\
\STATE $\theta \gets \text{OptimizerStep}(\theta, g^\top \frac{\partial W_b^{\prime}}{\partial \theta})$
\ENDFOR
\end{algorithmic}
\end{small}
\end{algorithm}
\end{minipage}
\vspace{-0.1in}
\end{wrapfigure}

Our design takes inspiration from attractor networks~\cite{attractor,localist}, where for each base
class one learns an ``attractor" that stores the relevant memory regarding that class. We call our
full model  ``dynamic attractors" as they may vary with each episode even after meta-learning. In
contrast if we only have the bias term $U_0$, i.e. a single attractor which is shared by all novel
classes, it will not change after meta-learning from one episode to the other. We call this model
variant the ``static attractor".

In summary, our meta parameters $\theta$ include $\phi$, $U_0$, $\gamma$ and $\tau$, which is on
 the same scale as as the number of paramters in $W_a$. It is important to note that $R(W_b,
 \theta)$ is convex w.r.t. $W_b$. Therefore, if we use the LR model as the classifier, the overall
 training objective on episodes in Eq. (\ref{eq:general_form}) is convex which implies that the
 optimum $W_b^*(\theta,S_b)$ is guaranteed to be unique and achievable. Here we emphasize that the
 optimal parameters $W_b^*$ are functions of parameters $\theta$ and few-shot samples $S_b$.

During meta-learning, $\theta$ are updated to minimize an expected loss of the query set $Q_{a+b}$
which contains both base and novel classes, averaging over all few-shot learning episodes,
\begin{equation}
\min_{\theta} ~~ \mathop{\mathbb{E}}_{\mathcal{E}} \left[ L^Q(\theta,S_b) \right] =
\mathop{\mathbb{E}}_{\mathcal{E}}\left[\sum_{j=1}^{M(K+K')} \sum_{c=1}^{K+K'} y_{j,c} \log
\hat{y}_{j,c}(\theta,S_b)\right],
\end{equation}
where the predicted class is
$
\hat{y}_{j}(\theta,S_b) = \softmax\left(\left[W_{a}^{\top} x_j, h \left( x_j;
W_{b}^{\ast}(\theta,S_b) \right) \right] \right)$.

\subsection{Learning via Recurrent Back-Propagation}
As there is no closed-form solution to the episodic objective (the optimization problem in Eq.
\ref{eq:general_form}), in each episode we need to minimize $L^S$ to obtain $W^*_b$ through an
iterative optimizer. The question is how to efficiently compute $\frac{\partial W_b^*}{\partial
\theta}$, \ie, back-propagating through the optimization. One option is to unroll the iterative
optimization process in the computation graph and use back-propagation through time (BPTT)
\cite{bptt}. However, the number of iterations for a gradient-based optimizer to converge can be on
the order of thousands, and BPTT can be computationally prohibitive. Another way is to use the
truncated BPTT \cite{tbptt} (T-BPTT) which optimizes for $T$ steps of gradient-based optimization,
and is commonly used in meta-learning problems. However, when $T$ is small the training objective
could be significantly biased.

Alternatively, the recurrent back-propagation (RBP) algorithm \cite{rbp2,rbp3,rbp} allows us to
back-propagate through the fixed point efficiently without unrolling the computation graph and
storing intermediate activations. Consider a vanilla gradient descent process on $W_b$ with step
size $\alpha$. The difference between two steps $\Phi$ can be written as $\Phi(W_b^{(t)}) =
W_b^{(t)} - F(W_b^{(t)})$, where $F(W_b^{(t)}) =W_b^{(t+1)} = W_b^{(t)} - \alpha \nabla
L^S(W_b^{(t)})$. Since $\Phi(W_b^{*}(\theta))$ is identically zero as a function of $\theta$,
using the implicit function theorem we have $\frac{\partial W_b^*}{\partial \theta} =
(I-J_{F,W_b^*}^\top)^{-1} \frac{\partial F}{\partial \theta}$, where $J_{F,W_b^*}$ denotes the
Jacobian matrix of the mapping $F$ evaluated at $W_b^*$. Algorithm~\ref{alg:energy} outlines the key
steps for learning the episodic objective using RBP in the incremental few-shot learning setting.
Note that the RBP algorithm implicitly inverts $(I-J^\top)$ by computing the matrix inverse vector
product, and has the same time complexity compared to truncated BPTT given the same number of
unrolled steps, but meanwhile RBP does not have to store intermediate activations.

\paragraph{Damped Neumann RBP}
To compute the matrix-inverse vector product $(I - J^\top)^{-1} v$, \cite{rbp} propose to use the
Neumann series: $(I-J^\top)^{-1}v  = \sum_{n=0}^\infty (J^\top)^{n}v \equiv \sum_{n=0}^\infty
v^{(n)}$. Note that $J^\top v$ can be computed by standard back-propagation. However, directly
applying the Neumann RBP algorithm sometimes leads to numerical instability. Therefore, we propose
to add a damping term $0 < \epsilon < 1$ to $I-J^\top$. This results in the following update:
$\tilde{v}^{(n)} = (J^\top - \epsilon I)^{n}v$. In practice, we found the damping term with
$\epsilon = 0.1$ helps alleviate the issue significantly.
% !TEX root = ../main.tex
\section{Experiments}
We experiment on two few-shot classification datasets, \textit{mini}-ImageNet and
\textit{tiered}-ImageNet. Both are subsets of ImageNet \cite{imagenet}, with images sizes reduced
to $84 \times 84$ pixels. We also modified the datasets to accommodate the
incremental few-shot learning settings.
\footnote{Code released at: \url{https://github.com/renmengye/inc-few-shot-attractor-public}}
\subsection{Datasets}
\begin{itemize}[leftmargin=*]
\item \textbf{\textit{mini}-ImageNet}
Proposed by \cite{matching}, \textit{mini}-ImageNet
contains 100 object classes and 60,000 images. We used the splits proposed by \cite{metalstm}, where
training, validation, and testing have 64, 16 and 20 classes respectively.
\item \textbf{\textit{tiered}-ImageNet}
Proposed by \cite{fewshotssl}, \textit{tiered}-ImageNet is a
larger subset of ILSVRC-12. It features a categorical split among training, validation, and testing
subsets. The categorical split means that classes that belong to the same high-level category, e.g.
“working dog” and "terrier" or some other dog breed, are not split between training, validation and
test. This is a harder task, but one that more strictly evaluates generalization to new classes. It
is also  an order of magnitude larger than \textit{mini}-ImageNet.
\end{itemize}

% !TEX root = ../main.tex
\begin{table}
\centering
\caption{Comparison of our proposed model with other methods}
\renewcommand{\arraystretch}{1.2}
\label{}
\begin{small}
% \begin{tabular}{l|p{2.8cm}|p{2.5cm}|p{2.8cm}}
\begin{tabular}{lp{3.2cm}p{2.5cm}p{3.8cm}}
\toprule
Method           & Few-shot learner     & Episodic objective      & Attention mechanism \\
% \hline\hline
\midrule
Imprint 
\cite{qi2018imprinting}         
                 & Prototypes          & N/A                     & N/A                \\
\hline
LwoF \cite{lwof} & Prototypes + base classes & N/A         & Attention on base classes      \\
\hline
Ours             & A fully trained classifier & Cross entropy on novel classes & Attention on learned attractors \\
\bottomrule
\end{tabular}
\end{small}
\end{table}
% !TEX root = ../main.tex
\begin{table}[t]
\vspace{-0.3in}
\begin{minipage}[t]{0.49\textwidth}
\begin{small}
\begin{center}
\caption{\textit{mini}-ImageNet 64+5-way results}
\vspace{-0.1in}
\label{tab:fewshot1}
\resizebox{\columnwidth}{!}{
\begin{tabular}{c|cc|cc}
\toprule
\multirow{2}{*}{Model} & \multicolumn{2}{c|}{1-shot}           & \multicolumn{2}{c}{5-shot} \\
          & Acc. $\uparrow$ & $\Delta \downarrow$ & Acc. $\uparrow$ & $\Delta \downarrow$ \\
\midrule
ProtoNet \cite{proto}
          & 42.73 $\pm$ 0.15      & -20.21      & 57.05 $\pm$ 0.10      & -31.72      \\
Imprint \cite{qi2018imprinting}
          & 41.10 $\pm$ 0.20      & -22.49      & 44.68 $\pm$ 0.23      & -27.68      \\
LwoF \cite{lwof}
          & 52.37 $\pm$ 0.20      & -13.65      & 59.90 $\pm$ 0.20      & -14.18      \\
Ours      & \tb{54.95} $\pm$ 0.30 & -11.84      & \tb{63.04} $\pm$ 0.30 & \tb{-10.66} \\
\bottomrule
\end{tabular}
}
\end{center}
\end{small}
\end{minipage}
\hfill
\begin{minipage}[t]{0.49\textwidth}
\begin{small}
\begin{center}
\caption{\textit{tiered}-ImageNet 200+5-way results}
\vspace{-0.1in}
\label{tab:fewshot2}
\resizebox{\columnwidth}{!}{
\begin{tabular}{c|cc|cc}
% \hline
\toprule
\multirow{2}{*}{Model} & \multicolumn{2}{c|}{1-shot} & \multicolumn{2}{c}{5-shot}  \\
 & Acc. $\uparrow$ & $\Delta \downarrow$ & Acc. $\uparrow$ & $\Delta \downarrow$    \\
% \hline\hline
\midrule
ProtoNet \cite{proto} & 30.04 $\pm$ 0.21 & -29.54 & 41.38 $\pm$ 0.28 & -26.39      \\
Imprint \cite{qi2018imprinting}
& 39.13 $\pm$ 0.15 & -22.26 & 53.60 $\pm$ 0.18 & -16.35 \\
LwoF \cite{lwof} & 52.40 $\pm$ 0.33 & -8.27  & 62.63 $\pm$ 0.31 & -6.72            \\
Ours        &  \tb{56.11} $\pm$ 0.33 & \tb{-6.11}  & \tb{65.52} $\pm$ 0.31 & \textbf{-4.48} \\
\bottomrule
\end{tabular}
}
\end{center}
\end{small}
\end{minipage}
\begin{center}
{\footnotesize $\Delta=$ average decrease in acc. caused by  \emph{joint} prediction within base and novel classes ($\Delta = \frac{1}{2} (\Delta_a+\Delta_b)$)\\
$\uparrow\, (\downarrow)$ represents higher (lower) is better.}
\end{center}
\vspace{-0.2in}
\end{table}

\subsection{Experiment setup}
We use a standard ResNet backbone \cite{resnet} to learn the feature representation through
supervised training. For \textit{mini}-ImageNet experiments, we follow \cite{mishra2017meta} and use
a modified version of ResNet-10. For \textit{tiered}-ImageNet, we use the standard ResNet-18
\cite{resnet}, but replace all batch normalization \cite{batchnorm} layers with group
normalization \cite{groupnorm}, as there is a large distributional shift from training to testing
in \textit{tiered}-ImageNet due to categorical splits. We used standard data augmentation, with
random crops and horizonal flips. We use the same pretrained checkpoint as the starting
point for meta-learning.

In the meta-learning stage as well as the final evaluation, we sample a few-shot episode from the
$\mathcal{D}_b$, together with a regular mini-batch from the $\mathcal{D}_a$. The base class images
are added to the query set of the few-shot episode. The base and novel classes are maintained in
equal proportion in our experiments. For all the experiments, we consider 5-way classification with
1 or 5 support examples (i.e. shots). In the experiments, we use a query set of size 25$\times$2
=50.

We use  L-BFGS~\cite{zhu1997algorithm} to solve the inner loop of our models to make sure $W_b$
converges. We use the ADAM~\cite{kingma2014adam} optimizer for meta-learning with a learning rate
of 1e-3, which decays by a factor of $10$ after 4,000 steps, for a total of 8,000 steps. We fix
recurrent backpropagation to 20 iterations and $\epsilon=0.1$.

We study two variants of the classifier network. The first is a logistic
regression model with a single weight matrix $W_b$. The second is a 2-layer fully connected MLP model
with 40 hidden units in the middle and $\tanh$ non-linearity. To make training more efficient, we
also add a shortcut connection in our MLP, which directly links the input to the output. In the
second stage of training, we keep all backbone weights frozen and only train the meta-parameters
$\theta$.

\subsection{Evaluation metrics}
We consider the following evaluation metrics:
1) overall accuracy on individual query sets and the joint query set (``Base'', ``Novel'', and
   ``Both''); and
2) decrease in performance caused by  \emph{joint} prediction within the base and novel classes,
   considered separately (``$\Delta_a$'' and ``$\Delta_b$''). Finally we take the average $\Delta =
   \frac{1}{2} (\Delta_a+\Delta_b)$ as a key measure of the overall decrease in accuracy.

\subsection{Comparisons}
We implemented and compared to three methods. First, we adapted Prototypical Networks \cite{proto}
to incremental few-shot settings.  For each base class we store a base representation, which is the
average representation (prototype) over all images belonging to the base class. During the few-shot
learning stage, we again average the representation of the few-shot classes and add them to the bank
of base representations. Finally, we retrieve the nearest neighbor by comparing the representation
of a test image with entries in the representation store. In summary, both $W_a$ and $W_b$ are
stored as the average representation of all images seen so far that belong to a certain class. We
also compare to the following methods:
\begin{itemize}[leftmargin=*]
    \item \textbf{Weights Imprinting (``Imprint'')} \cite{qi2018imprinting}: the base weights $W_a$
are learned regularly through supervised pre-training, and $W_b$ are computed using prototypical
averaging.
    \item \textbf{Learning without Forgetting (``LwoF'')} \cite{lwof}: Similar to
\cite{qi2018imprinting}, $W_b$ are computed using prototypical averaging. In addition, $W_a$ is
finetuned during episodic meta-learning. We implemented the most advanced variants proposed in the
paper, which involves a class-wise attention mechanism. This model is the previous state-of-the-art
method on incremental few-shot learning, and has better performance compared to other low-shot models
\cite{wang2018lowshot,hariharan2017lowshot}.
\end{itemize}
\subsection{Results}
We first evaluate our vanilla approach on the standard few-shot classification benchmark where no
base classes are present in the query set. Our vanilla model consists of a pretrained CNN and a
single-layer logistic regression with weight decay learned from scratch; this model performs on-par
with other competitive meta-learning approaches (1-shot 55.40 $\pm$ 0.51, 5-shot 70.17 $\pm$ 0.46).
Note that our model uses the same backbone architecture as \cite{mishra2017meta} and \cite{lwof},
and is directly comparable with their results. Similar findings of strong results using simple
logistic regression on few-shot classification benchmarks are also recently reported in
\cite{closerlook}. Our full model has similar performance as the vanilla model on pure few-shot
benchmarks, and the full table is available in Supp. Materials.

Next, we compare our models to other methods on incremental few-shot learning benchmarks in
Tables~\ref{tab:fewshot1} and  \ref{tab:fewshot2}. On both benchmarks, our best performing model
shows a significant margin over the prior works that predict the prototype representation without
using an iterative optimization \cite{proto,qi2018imprinting,lwof}.

% !TEX root = ../main.tex
\begin{table}[t]
\vspace{-0.3in}
\begin{minipage}[t]{0.49\textwidth}
\begin{small}
\begin{center}
\caption{Ablation studies on {\it mini}-ImageNet}
\vspace{-0.1in}
\label{tab:mini-ablation}
\resizebox{\columnwidth}{!}{
\begin{tabular}{c|cc|cc}
\toprule
          & \multicolumn{2}{c|}{1-shot}        & \multicolumn{2}{c}{5-shot} \\
          & Acc. $\ua$            & \D         & Acc. $\ua$            & \D          \\    
\midrule
LR        & 52.74 $\pm$ 0.24      & -13.95     & 60.34 $\pm$ 0.20      & -13.60      \\
LR +S     & 53.63 $\pm$ 0.30      & -12.53     & 62.50 $\pm$ 0.30      & -11.29      \\
LR +A     & \tb{55.31} $\pm$ 0.32 & \tb{-11.72}& 63.00 $\pm$ 0.29      & -10.80      \\
\midrule                                                                                        
MLP       & 49.36 $\pm$ 0.29      & -16.78     & 60.85 $\pm$ 0.29      & -12.62      \\
MLP +S    & 54.46 $\pm$ 0.31      & -11.74     & 62.79 $\pm$ 0.31      & -10.77      \\
MLP +A    & 54.95 $\pm$ 0.30      & -11.84     & \tb{63.04} $\pm$ 0.30 & \tb{-10.66} \\
\bottomrule
\end{tabular}
}
\end{center}
\end{small}
\end{minipage}
\hfill
\begin{minipage}[t]{0.49\textwidth}
\begin{small}
\begin{center}
\caption{Ablation studies on {\it tiered}-ImageNet}
\vspace{-0.1in}
\label{tab:tiered-ablation}
\resizebox{\textwidth}{!}{
\begin{tabular}{c|cc|cc}
\toprule
          & \multicolumn{2}{c|}{1-shot}        & \multicolumn{2}{c}{5-shot} \\
          & Acc. $\ua$            & \D         & Acc. $\ua$            & \D          \\
\midrule                                                                          
LR        & 48.84 $\pm$ 0.23      & -10.44     & 62.08 $\pm$ 0.20      & -8.00       \\
LR +S     & 55.36 $\pm$ 0.32      & -6.88      & 65.53 $\pm$ 0.30      & -4.68       \\
LR +A     & 55.98 $\pm$ 0.32      & \tb{-6.07} & 65.58 $\pm$ 0.29      & \tb{-4.39}  \\
\midrule                                                                                                       
MLP       & 41.22 $\pm$ 0.35      & -10.61     & 62.70 $\pm$ 0.31      & -7.44       \\
MLP +S    & \tb{56.16} $\pm$ 0.32 & -6.28      & \tb{65.80} $\pm$ 0.31 & -4.58       \\
MLP +A    & 56.11 $\pm$ 0.33      & 6.11       & 65.52 $\pm$ 0.31      & -4.48       \\
\bottomrule
\end{tabular}
}
\end{center}
\end{small}
\end{minipage}
\begin{center}
{\footnotesize ``+S'' stands for static attractors, and ``+A'' for attention attractors.}
\end{center}
\end{table}

\subsection{Ablation studies}
To understand the effectiveness of each part of the proposed model, we consider the following
variants:
\begin{itemize}[leftmargin=*]
\item \textbf{Vanilla (``LR, MLP'')} optimizes a logistic regression or an MLP network at each
few-shot episode, with a weight decay regularizer.
\item \textbf{Static attractor (``+S'')} learns a fixed attractor center $u$ and attractor
slope $\gamma$ for all classes. 
\item \textbf{Attention attractor (``+A'')} learns the full attention attractor model.
For MLP models, the weights below the final layer are controlled by attractors predicted
by the average representation across all the episodes. $f_\phi$ is an MLP with one hidden layer of
50 units.
\end{itemize}
Tables~\ref{tab:mini-ablation} and \ref{tab:tiered-ablation} shows the ablation experiment results.
In all cases, the learned regularization function shows better performance than a manually set
weight decay constant on the classifier network, in terms of both jointly predicting base and novel
classes, as well as less degradation from individual prediction. On \textit{mini}-ImageNet, our
attention attractors have a clear advantage over static attractors.

Formulating the classifier as an MLP network is slightly better than the linear models in our
experiments. Although the final performance is similar, our RBP-based algorithm have the flexibility
of adding the fast episodic model with more capacity. Unlike \cite{diffsolver}, we do not rely on an
analytic form of the gradients of the optimization process.

% !TEX root = ../main.tex
\begin{figure}[t]
\label{fig:bptt}
\begin{minipage}[c]{\textwidth}
\centering
\includegraphics[width=0.49\textwidth,trim={0cm 0cm 0cm 0cm},clip]{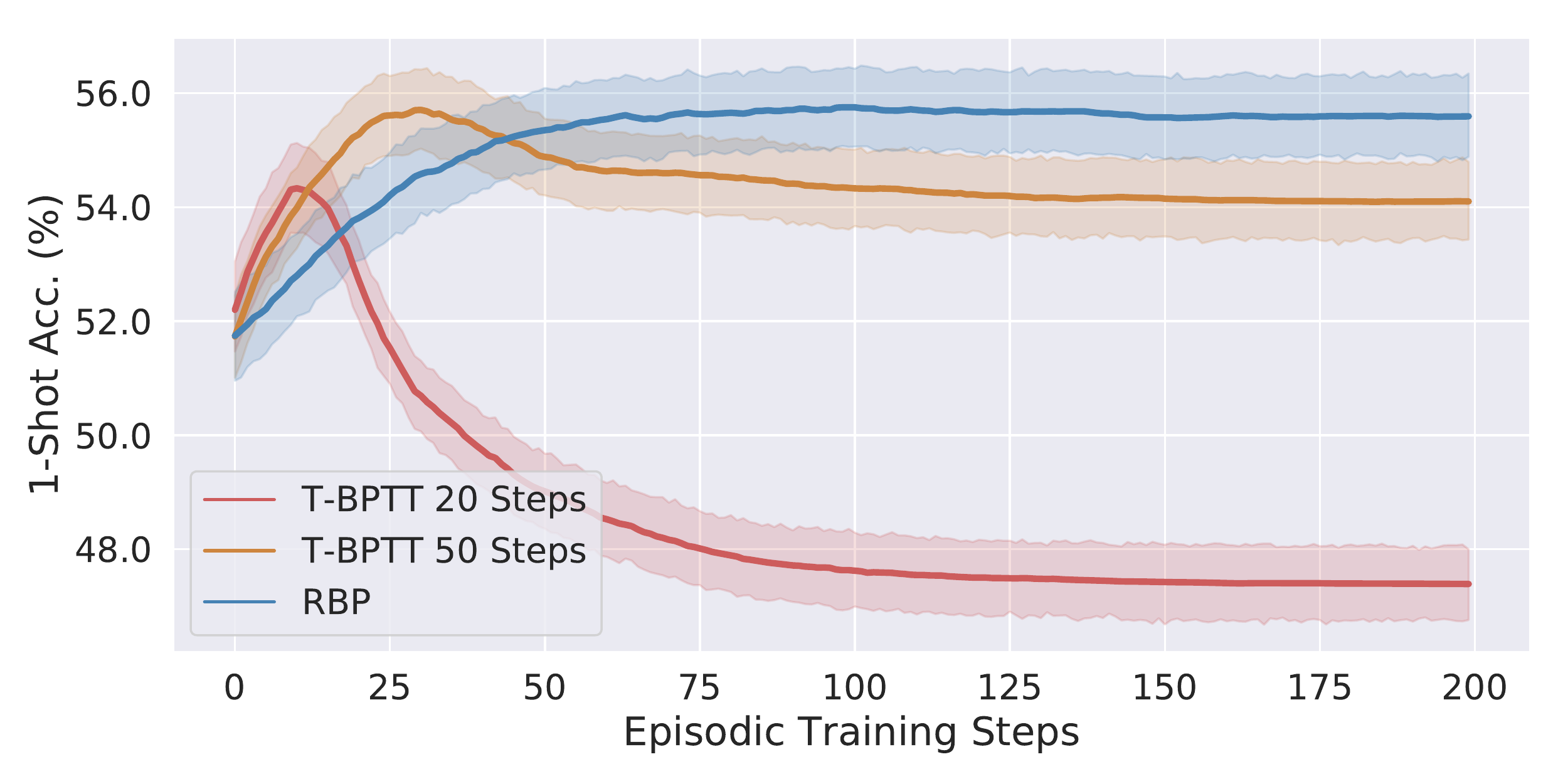}
\hfill
\includegraphics[width=0.49\textwidth,trim={0cm 0cm 0cm 0cm},clip]{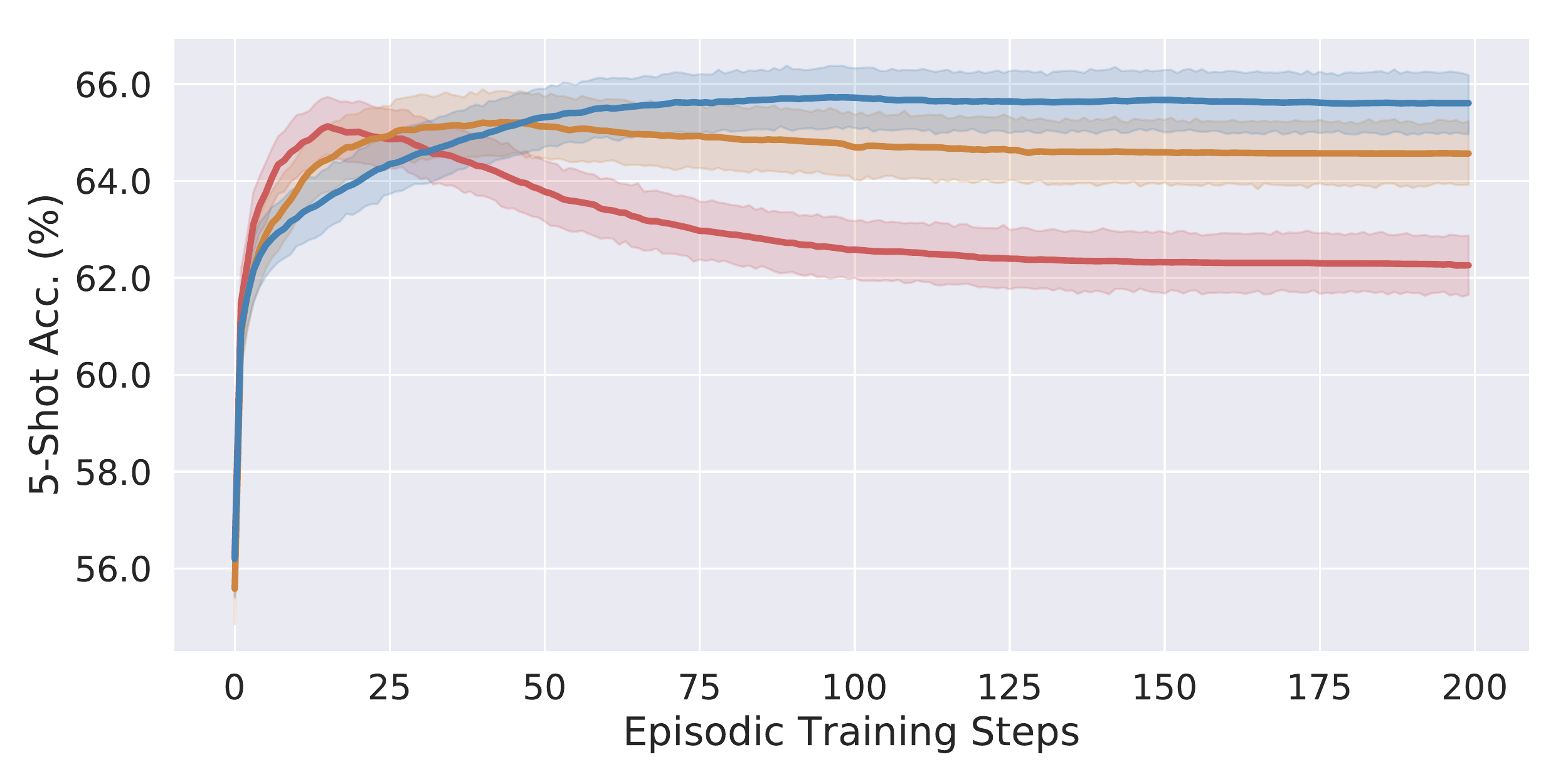}
\end{minipage}
\caption{Learning the proposed model using truncated BPTT vs. RBP. Models
are evaluated with 1-shot (left) and 5-shot (right) 64+5-way episodes,
with various number of gradient descent steps.}
\label{fig:bptt}
\end{figure}
% !TEX root = ../main.tex
\begin{figure}[t]
\vspace{-0.2in}
\centering
\begin{minipage}[c]{\textwidth}
\centering
\begin{small}
\begin{tabular}{cc}
\includegraphics[width=0.47\textwidth,trim={2.8cm 1cm 2.5cm 1cm},clip]{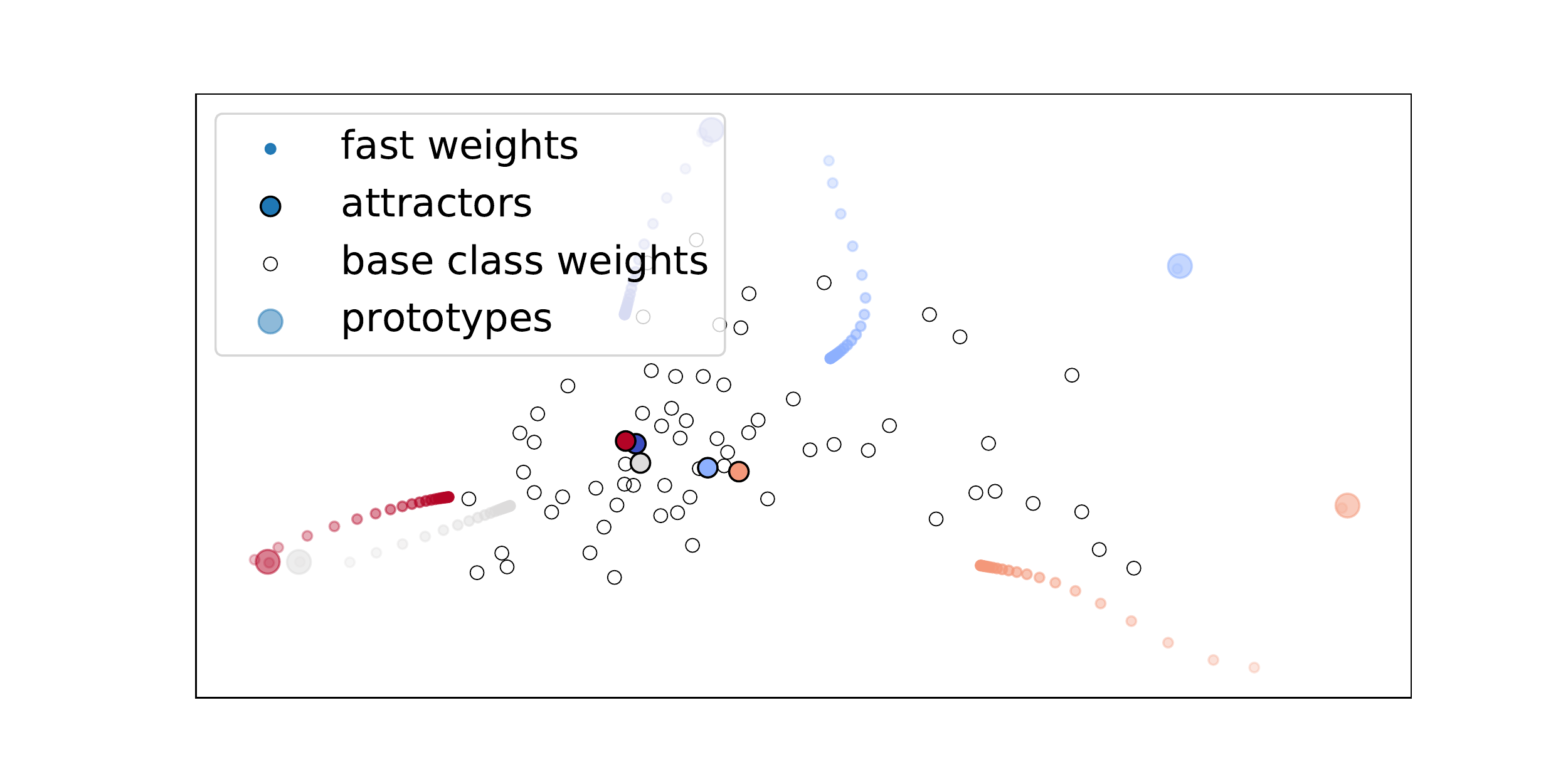} & 
\includegraphics[width=0.47\textwidth,trim={2.8cm 1cm 2.5cm 1cm},clip]{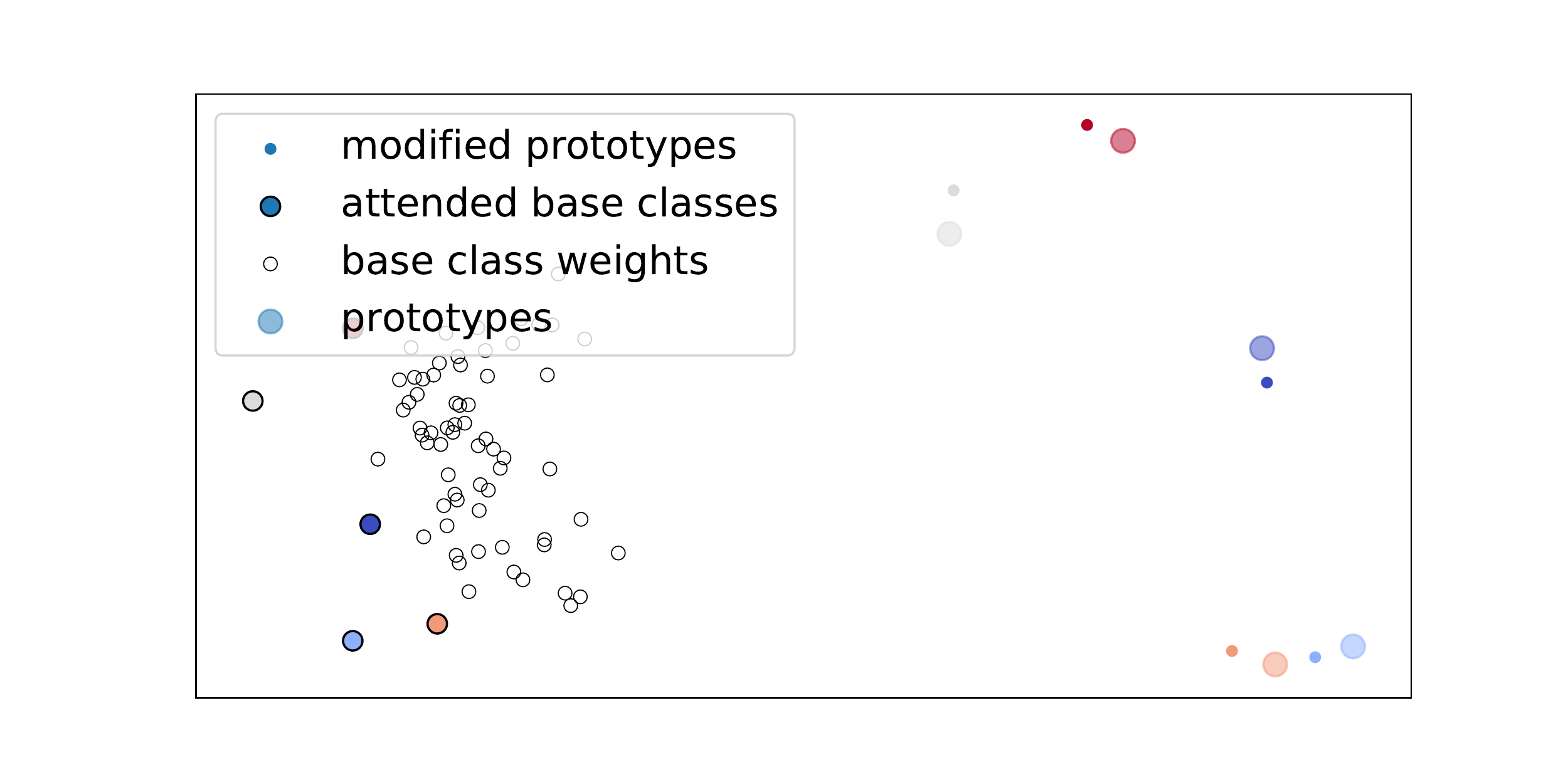}\\
(a) Ours & (b) LwoF \cite{lwof}
\end{tabular}
\end{small}
\end{minipage}
\caption{Visualization of a 5-shot 64+5-way episode using PCA. 
\textbf{Left:} Our attractor model learns to
``pull'' prototypes (large colored circles) towards base class weights (white circles). We visualize the trajectories during episodic training; \textbf{Right:} Dynamic few-shot learning without
forgetting \cite{lwof}.}
\label{fig:vizproc}
\vspace{-0.1in}
\end{figure}
% !TEX root = ../main.tex
\begin{wrapfigure}{R}{0.5\textwidth}
\vspace{-0.1in}
\centering
\hfill
\includegraphics[width=0.5\textwidth,trim={0cm 0cm 0cm 0cm},clip]{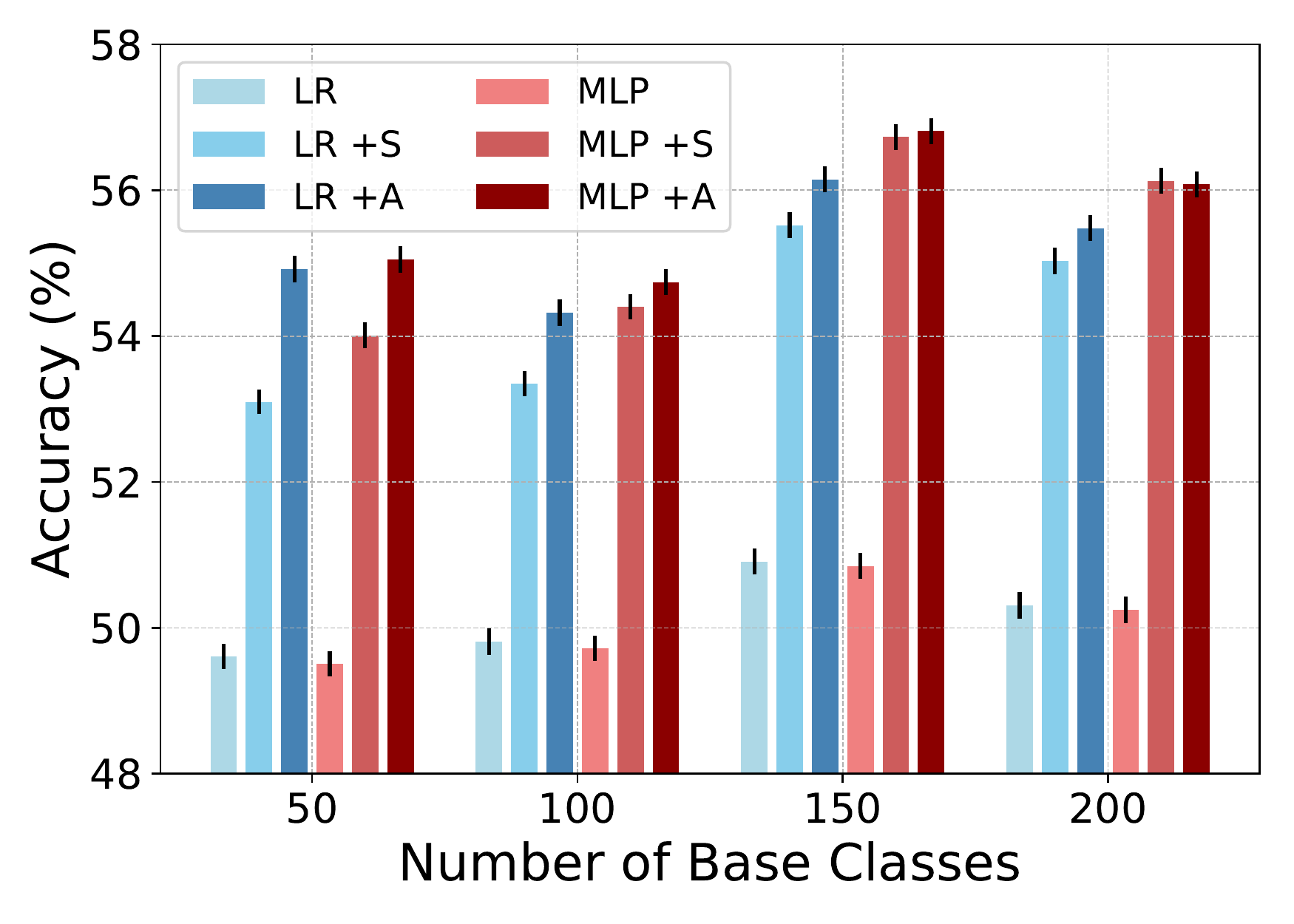}
\caption{Results on \textit{tiered}-ImageNet with \{50, 100, 150, 200\} base classes.
}
\label{fig:snapshot}
\vspace{-0.25in}
\end{wrapfigure}
\paragraph{Comparison to truncated BPTT (T-BPTT)}
An alternative way to learn the regularizer is to unroll the inner optimization for a fixed number
of steps in a differentiable computation graph, and then back-propagate through time. Truncated BPTT
is a popular learning algorithm in many recent meta-learning approaches
\cite{gd2,metalstm,maml,mbpa,metareg}. Shown in Figure~\ref{fig:bptt}, the performance of T-BPTT
learned models are comparable to ours; however, when solved to convergence at test time, the
performance of T-BPTT models drops significantly. This is expected as they are only guaranteed to
work well for a certain number of steps, and failed to learn a good regularizer.
While an early-stopped T-BPTT model can do equally well, in practice it is hard to tell when to stop; whereas for the RBP model, doing the full episodic training is very fast since the number of support examples is small. 

\paragraph{Visualization of attractor dynamics} We visualize attractor dynamics in
Figure~\ref{fig:vizproc}. Our learned attractors pulled the fast weights close towards the base
class weights. In comparison, \cite{lwof} only modifies the prototypes slightly.

\paragraph{Varying the number of base classes}
While the framework proposed in this paper cannot be directly applied on class-incremental continual
learning, as there is no module for memory consolidation, we can simulate the continual learning
process by varying the number of base classes, to see how the proposed models are affected by
different stages of continual learning. Figure~\ref{fig:snapshot} shows that the learned
regularizers consistently improve over baselines with weight decay only. The overall accuracy
increases from 50 to 150 classes due to better representations on the backbone network, and drops
at 200 classes due to a more challenging classification task.
% !TEX root = ../main.tex
% \vspace{-0.1in}
\section{Conclusion}
% \vspace{-0.1in}
Incremental few-shot learning, the ability to jointly predict based on a set of pre-defined concepts
as well as additional novel concepts, is an important step towards making machine learning models
more flexible and usable in everyday life. In this work, we propose an attention attractor model,
which regulates a per-episode training objective by attending to the set of base classes. We show
that our iterative model that solves the few-shot objective till convergence is better than
baselines that do one-step inference, and that recurrent back-propagation is an effective and
modular tool for learning in a general meta-learning setting, whereas truncated back-propagation
through time fails to learn functions that converge well. %During evaluation of few-shot episodes,
% our attention attractor network learns to remember the base classes without needing to review
% examples from the original training set. 
Future directions of this work include sequential iterative
learning of few-shot novel concepts, and hierarchical memory organization.
% !TEX root = ../main.tex
\paragraph{Acknowledgment} Supported by NSERC and the Intelligence Advanced Research Projects
Activity (IARPA) via Department of Interior/Interior Business Center (DoI/IBC) contract number
D16PC00003. The U.S. Government is authorized to reproduce and distribute reprints for Governmental
purposes notwithstanding any copyright annotation thereon. Disclaimer: The views and conclusions
contained herein are those of the authors and should not be interpreted as necessarily representing
the official policies or endorsements, either expressed or implied, of IARPA, DoI/IBC, or the U.S.
Government.
{
\setstretch{0.93}
\bibliography{ref}
\bibliographystyle{abbrv}
}
\setstretch{1.0}
\appendix
\if\arxiv1
% !TEX root = ../main.tex
\section{Regular Few-Shot Classification}
We include standard 5-way few-shot classification results in Table~\ref{tab:fewshot1_novel}. As
mentioned in the main text, a simple logistic regression model can achieve competitive performance
on few-shot classification using pretrained features. Our full model shows similar performance on
regular few-shot classification. This confirms that the learned regularizer is mainly solving the
interference problem between the base and novel classes.
\vspace{0.1in}

\begin{table}[h!]
\begin{small}
\begin{center}
\caption{Regular 5-way few-shot classification on \textit{mini-ImageNet}.
Note that this is purely few-shot, with no base classes. Applying logistic regression on pretrained
features achieves performance on-par with other competitive meta-learning approaches. * denotes our
own implementation.}
\label{tab:fewshot1_novel}
% \begin{minipage}[c]{0.5\textwidth}
% \hfill
% \resizebox{\columnwidth}{!}{
% \begin{tabular}{|c|c|c|c|}
\begin{tabular}{cccc}
% \hline
\toprule
Model        & Backbone & 1-shot                & 5-shot                \\
% \hline\hline 
\midrule                                                             
MatchingNets \cite{matching} 
             & C64      & 43.60                 & 55.30                 \\
Meta-LSTM \cite{metalstm} 
           & C32      & 43.40 $\pm$ 0.77      & 60.20 $\pm$ 0.71      \\
MAML \cite{maml}
             & C64      & 48.70 $\pm$ 1.84      & 63.10 $\pm$ 0.92      \\
RelationNet \cite{relationnet} 
             & C64      & 50.44 $\pm$ 0.82      & 65.32 $\pm$ 0.70      \\
R2-D2  \cite{diffsolver} 
           & C256     & 51.20 $\pm$ 0.60      & 68.20 $\pm$ 0.60      \\
SNAIL \cite{mishra2017meta} 
           & ResNet   & 55.71 $\pm$ 0.99      & 68.88 $\pm$ 0.92      \\
ProtoNet \cite{proto} 
             & C64      & 49.42 $\pm$ 0.78      & 68.20 $\pm$ 0.66      \\
ProtoNet*  \cite{proto} 
             & ResNet   & 50.09 $\pm$ 0.41      & 70.76 $\pm$ 0.19      \\
LwoF \cite{lwof} 
             & ResNet   & 55.45 $\pm$ 0.89      & \tb{70.92} $\pm$ 0.35 \\
% LwoF*        & ResNet   & \textbf{56.97} $\pm$ 0.24 & 70.50 $\pm$ 0.36 \\
% \hline
\midrule
LR           & ResNet   & 55.40 $\pm$ 0.51      & 70.17 $\pm$ 0.46 \\
% LR +S        & ResNet   & 55.06 $\pm$ 0.52      & 70.32 $\pm$ 0.46 \\
Ours Full     & ResNet   & \tb{55.75} $\pm$ 0.51 & 70.14 $\pm$ 0.44 \\
% \hline
\bottomrule
\end{tabular}
% }
% \end{minipage}
\end{center}
\end{small}
\end{table}

\section{Visualization of Few-Shot Episodes}
We include more visualization of few-shot episodes in Figure~\ref{fig:moreviz}, highlighting the differences between our method and ``Dynamic Few-Shot Learning without Forgetting''~\cite{lwof}.
\begin{figure}[h!]
\centering
\begin{minipage}[c]{\textwidth}
\begin{small}
\begin{tabular}{cc}
\includegraphics[width=0.45\textwidth,trim={2.8cm 1cm 2.5cm 1cm},clip]{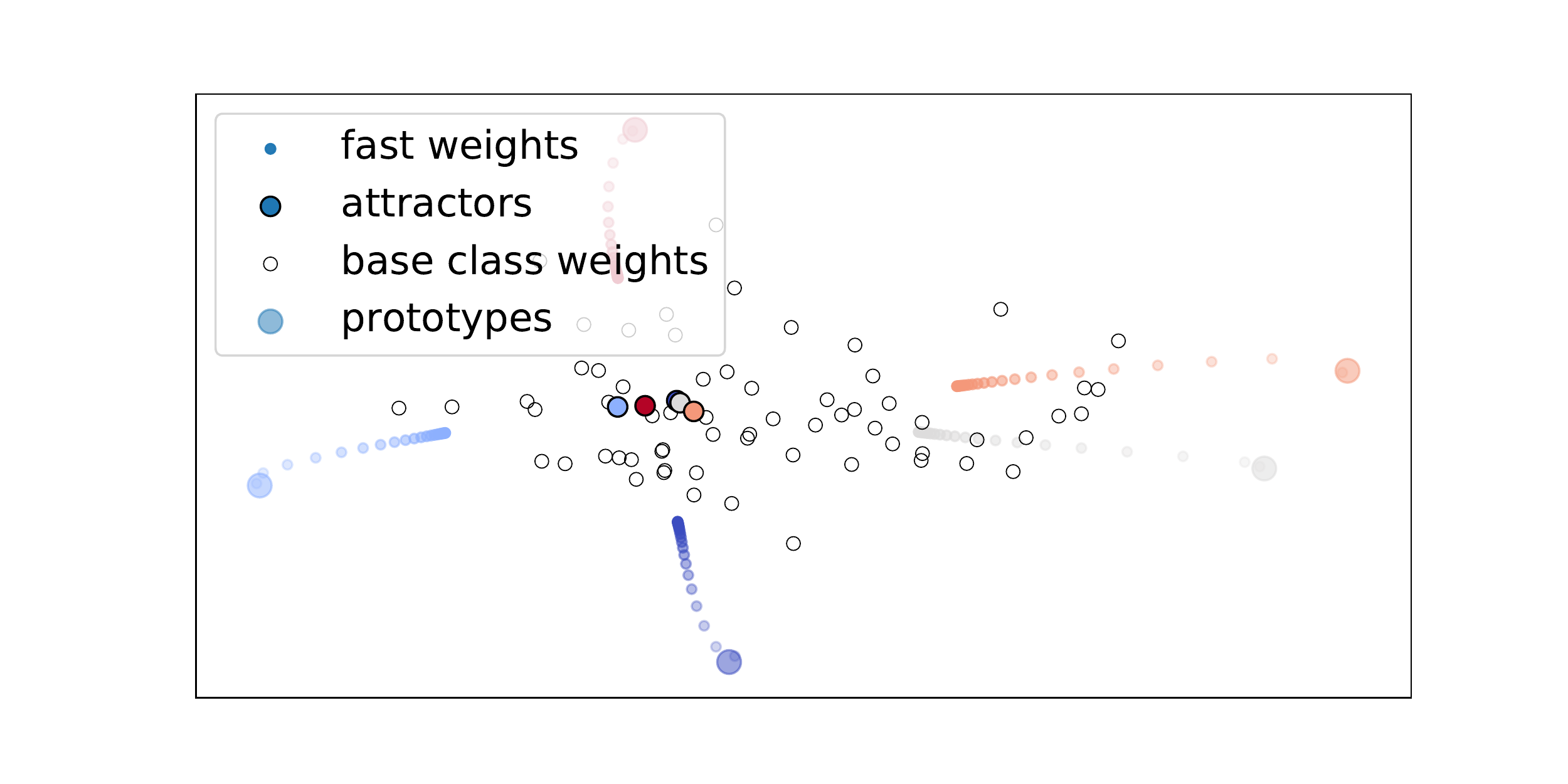} & 
\includegraphics[width=0.45\textwidth,trim={2.8cm 1cm 2.5cm 1cm},clip]{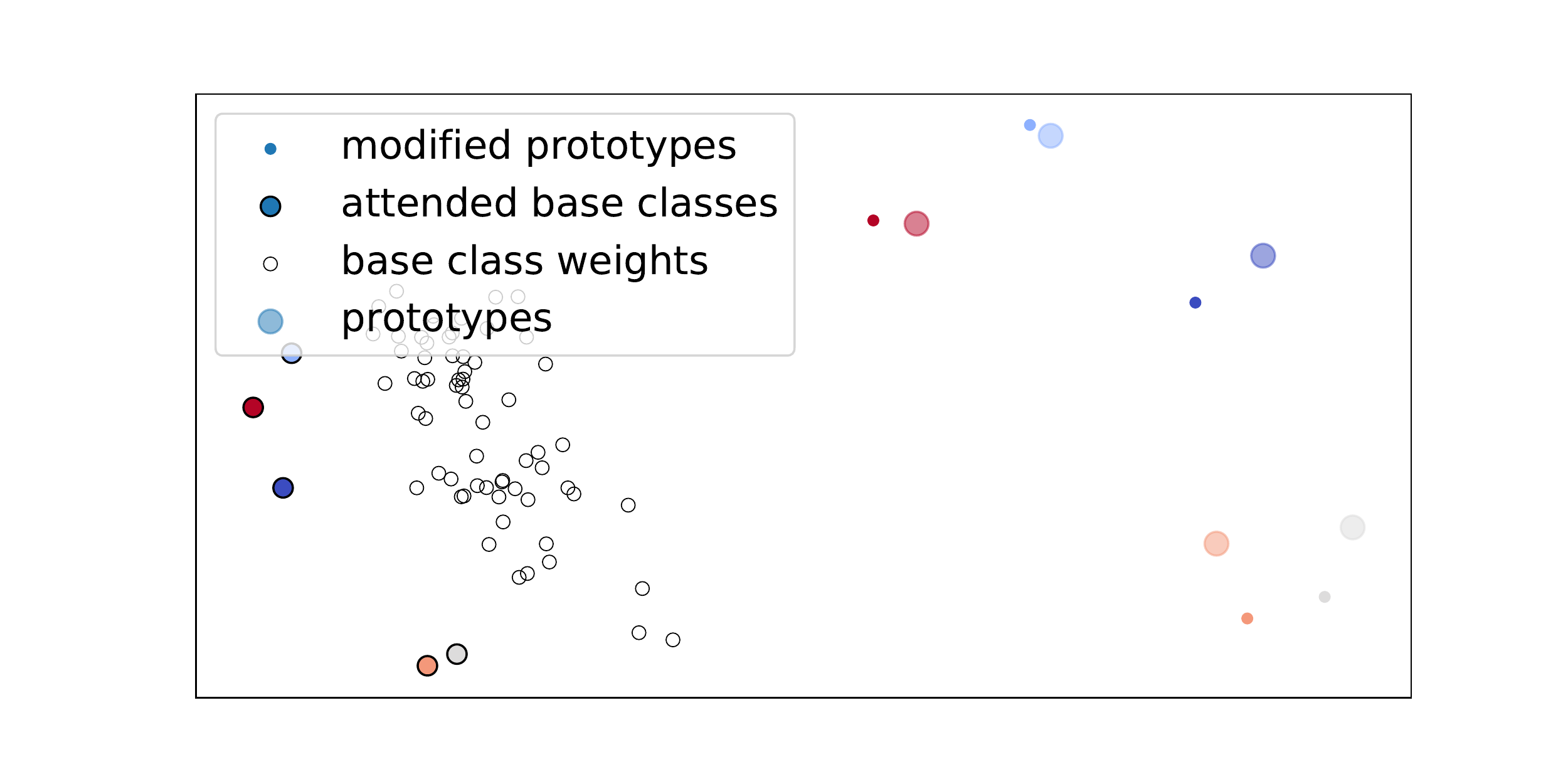}\\
\hline
\includegraphics[width=0.45\textwidth,trim={2.8cm 1cm 2.5cm 1cm},clip]{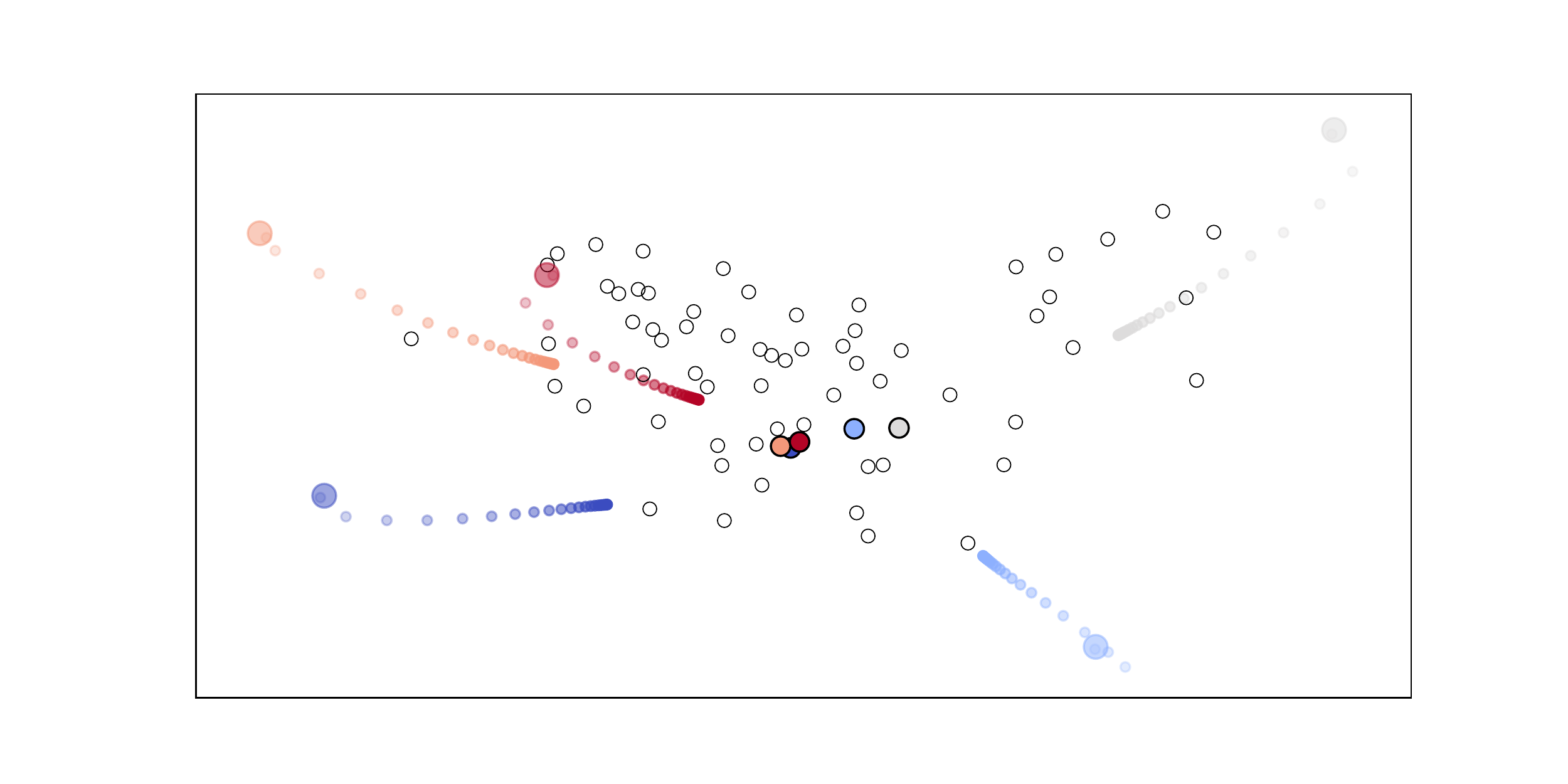} & 
\includegraphics[width=0.45\textwidth,trim={2.8cm 1cm 2.5cm 1cm},clip]{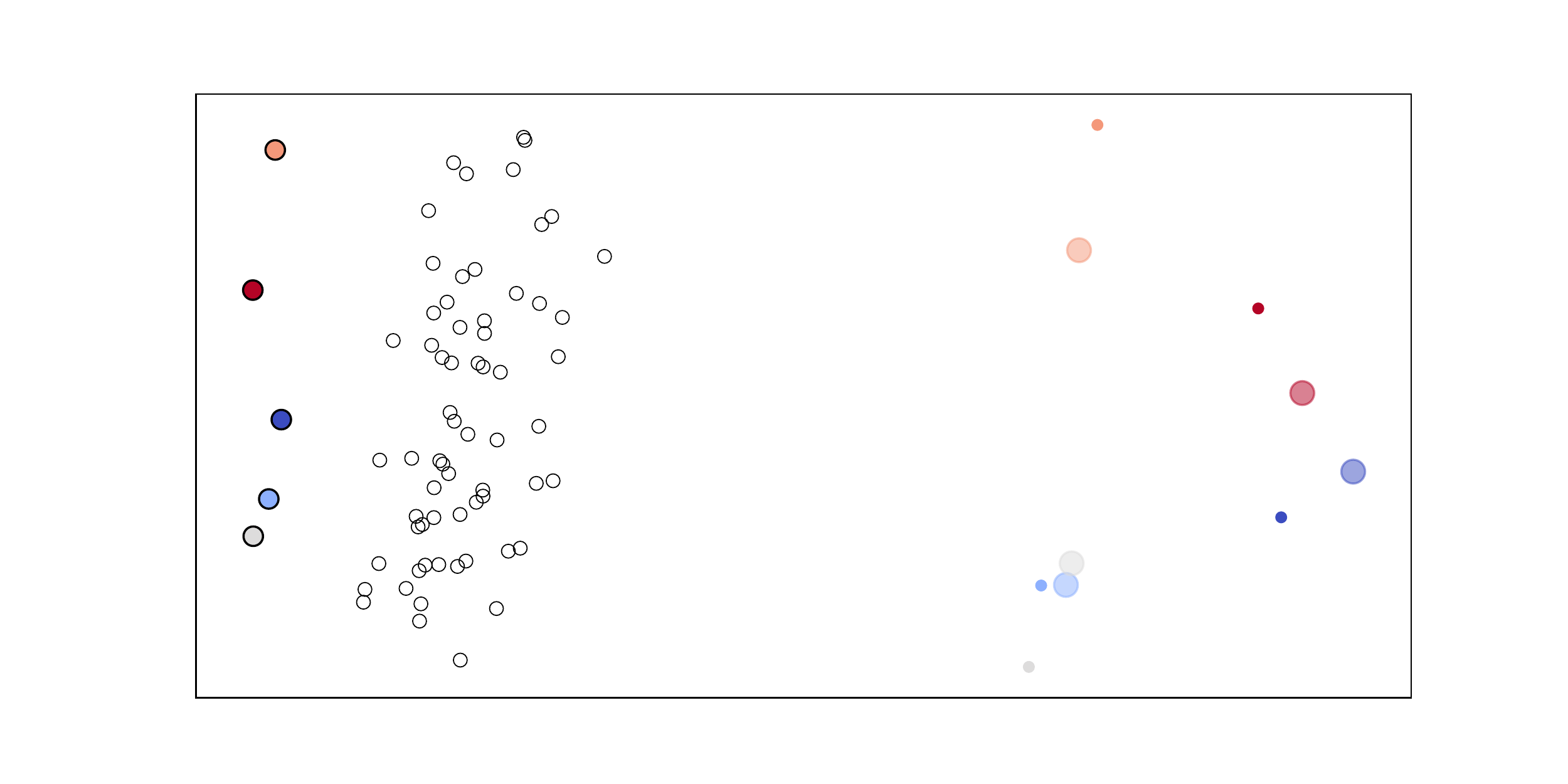}\\
\hline
\includegraphics[width=0.45\textwidth,trim={2.8cm 1cm 2.5cm 1cm},clip]{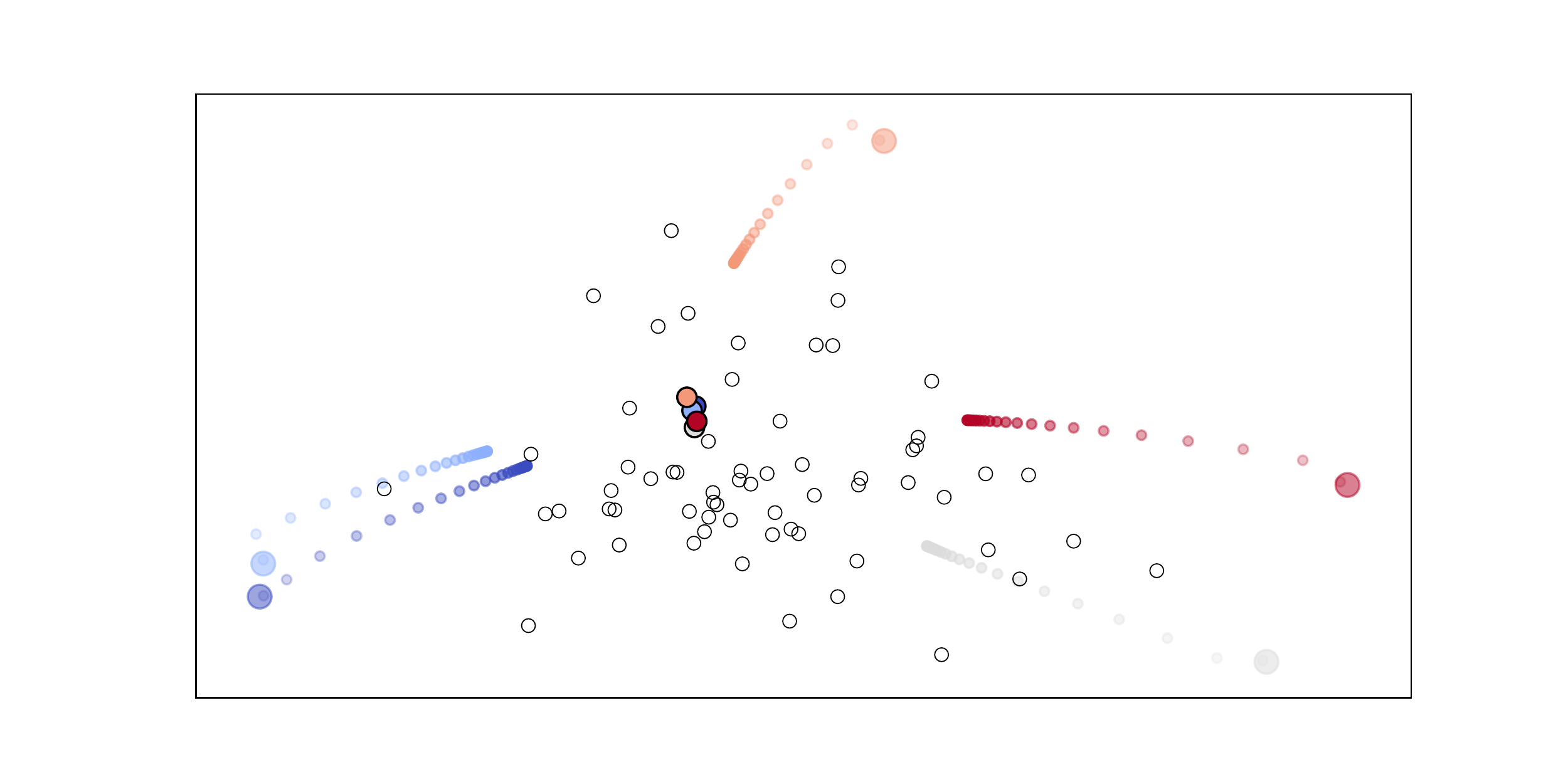} & 
\includegraphics[width=0.45\textwidth,trim={2.8cm 1cm 2.5cm 1cm},clip]{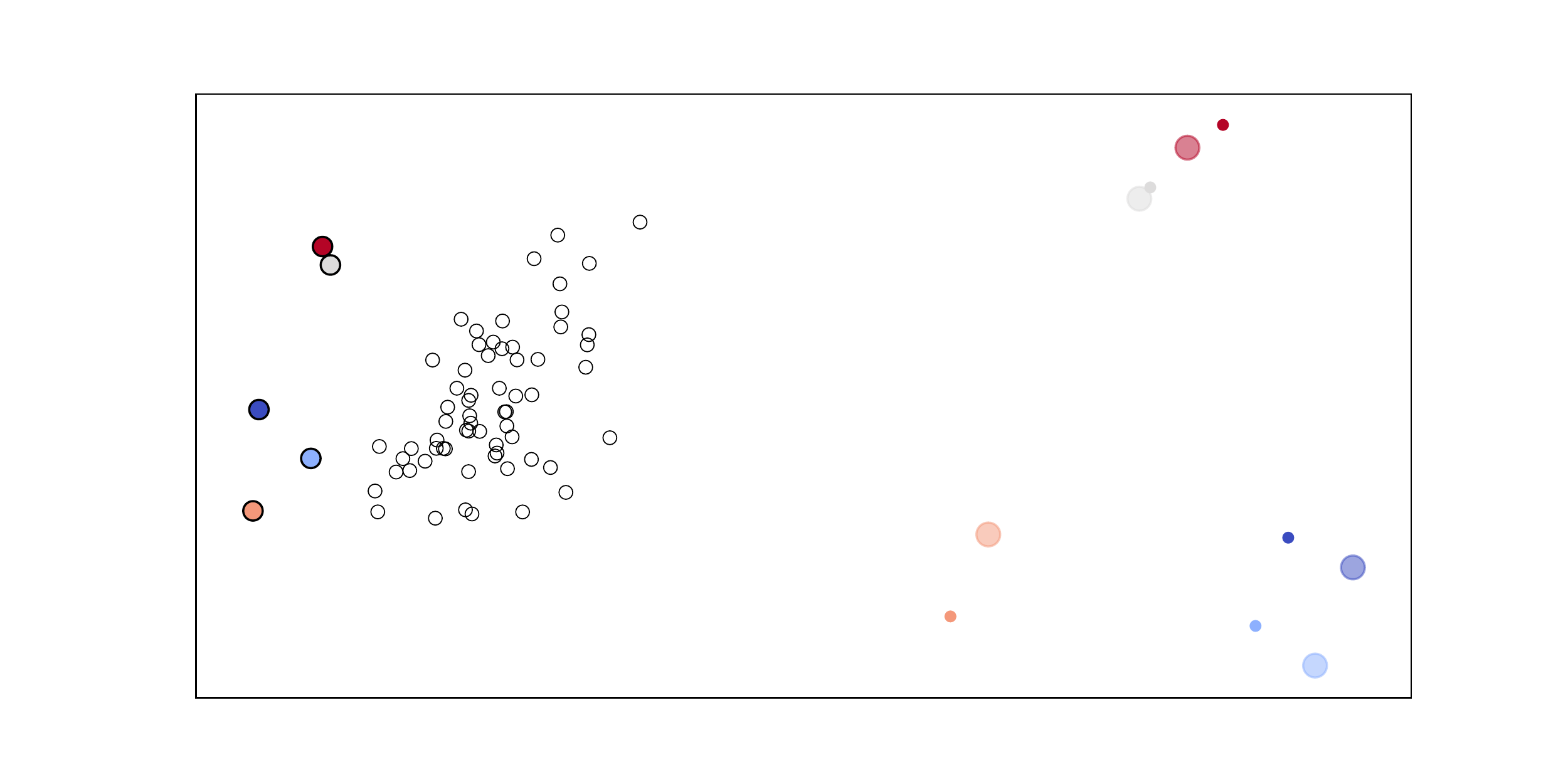}\\
(a) Ours & (b) LwoF \cite{lwof}
\end{tabular}
\end{small}
\end{minipage}
\caption{Visualization of 5-shot 64+5-way episodes on \textit{mini}-ImageNet using PCA.}
\label{fig:moreviz}
\end{figure}

\section{Visualization of Attention Attractors}
To further understand the attractor mechanism, we picked 5 semantic classes in
\textit{mini}-ImageNet and visualized their the attention attractors across 20 episodes, shown in
Figure~\ref{fig:attractorviz}. The attractors roughly form semantic clusters, whereas the static
attractor stays in the center of all attractors.
% !TEX root = ../supp.tex
\begin{figure}[h!]
\centering
\includegraphics[width=0.6\textwidth,trim={1cm 0.5cm 1.6cm 1cm},clip]{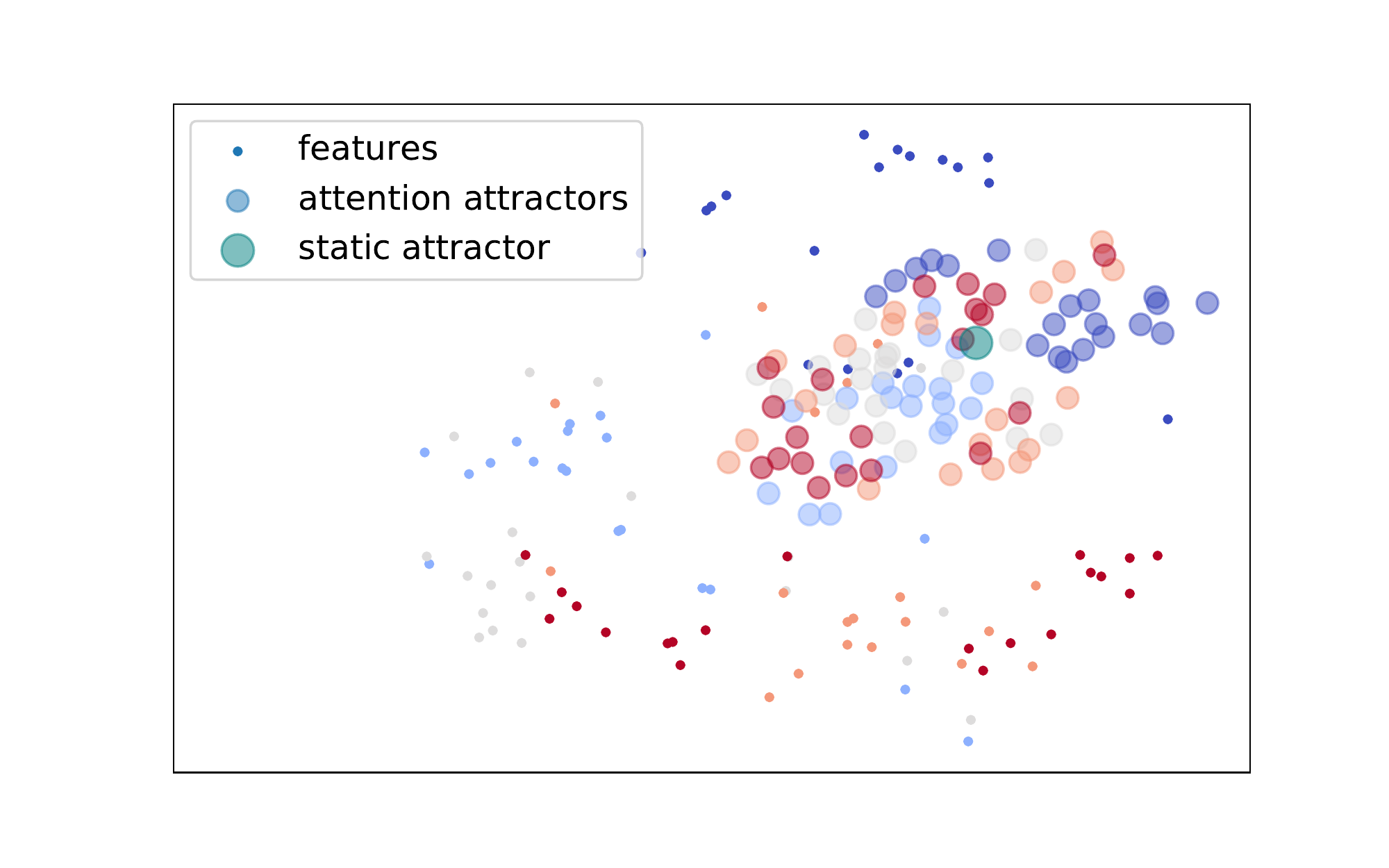}
\caption{Visualization of example features and attractors using t-SNE. This plot shows a 5-way
5-shot episode on \textit{mini}-ImageNet. 512-dimensional feature vectors and attractor vectors
are projected to a 2-dim space. Color represents the label class of the example. The static
attractor (\textcolor{teal}{teal}) appears at the center of the attention attractors, which roughly
form clusters based on the classes.}
\label{fig:attractorviz}
\end{figure}

\begin{table}
\centering
\caption{Full ablation results on 64+5-way {\it mini}-ImageNet}
\resizebox{0.8\textwidth}{!}{
\begin{tabular}{c|cccc|cccc}
\toprule
          & \multicolumn{4}{c|}{1-shot}           & \multicolumn{4}{c}{5-shot} \\
          & Acc. $\ua$            & \D         & \Da    &\Db     & Acc. $\ua$            & \D         & \Da    & \Db     \\
\midrule                                                                                                                
LR        & 52.74 $\pm$ 0.24      & -13.95     & -8.98  & -24.32 & 60.34 $\pm$ 0.20      & -13.60     & -10.81 & -15.97  \\
LR +S     & 53.63 $\pm$ 0.30      & -12.53     & -9.44  & -15.62 & 62.50 $\pm$ 0.30      & -11.29     & -13.84 & -8.75   \\
LR +A     & \tb{55.31} $\pm$ 0.32 & \tb{-11.72}& -12.72 & -10.71  & 63.00 $\pm$ 0.29      & -10.80     & -13.59 & -8.01   \\
\midrule                                                                                                                             
MLP       & 49.36 $\pm$ 0.29      & -16.78     & -8.95  & -24.61 & 60.85 $\pm$ 0.29      & -12.62     & -11.35 & -13.89  \\
MLP +S    & 54.46 $\pm$ 0.31      & -11.74     & -12.73 & -10.74 & 62.79 $\pm$ 0.31      & -10.77     & -12.61 & -8.80   \\
MLP +A    & 54.95 $\pm$ 0.30      & -11.84     & -12.81 & -10.87 & \tb{63.04} $\pm$ 0.30 & \tb{-10.66}& -12.55 & -8.77   \\
\bottomrule
\end{tabular}
}
\end{table}

\begin{table*}[t!]
\centering
\caption{Full ablation results on 200+5-way {\it tiered}-ImageNet}
\resizebox{0.8\textwidth}{!}{
\begin{tabular}{c|cccc|cccc}
\toprule
          & \multicolumn{4}{c|}{1-shot}           & \multicolumn{4}{c}{5-shot} \\
          & Acc. $\ua$            & \D         & \Da    & \Db    & Acc. $\ua$            & \D         & \Da    & \Db     \\
\midrule                                                                                                              
LR        & 48.84 $\pm$ 0.23      & -10.44     & -11.65 & -9.24  & 62.08 $\pm$ 0.20      & -8.00      & -5.49  & -10.51  \\
LR +S     & 55.36 $\pm$ 0.32      & -6.88      & -7.21  & -6.55  & 65.53 $\pm$ 0.30      & -4.68      & -4.72  & -4.63   \\
LR +A     & 55.98 $\pm$ 0.32      & \tb{-6.07} & -6.64  & -5.51  & 65.58 $\pm$ 0.29      & \tb{-4.39} & -4.87  & -3.91   \\
\midrule                                                                                                                                           
MLP       & 41.22 $\pm$ 0.35      & -10.61     & -11.25 & -9.98  & 62.70 $\pm$ 0.31      & -7.44      & -6.05  & -8.82   \\
MLP +S    & \tb{56.16} $\pm$ 0.32 & -6.28      & -6.83  & -5.73  & \tb{65.80} $\pm$ 0.31 & -4.58      & -4.66  & -4.51   \\
MLP +A    & 56.11 $\pm$ 0.33      & 6.11       & -6.79  & -5.43  & 65.52 $\pm$ 0.31      & -4.48      & -4.91  & -4.05   \\
\bottomrule
\end{tabular}
}
\end{table*}

\section{Dataset Statistics}
In this section, we include more details on the datasets we used in our experiments.
% We include the dataset statistics in Table~\ref{tab:stats}. In \textit{mini}-ImageNet, we use the
% training set for both pretraining and meta-learning. For testing base class classification
% performance, we included the same val/test set as \cite{lwof}. Since the meta-training set is same
% as meta-learning, in each training episode, we masked out the 5 base classes in the base classifier,
% to ``pretend'' they are few-shot classes. In \textit{tiered}-ImageNet, we splits the original
% training set, Train-A and Train-B, for pretraining and meta-learning respectively.

\begin{table}[h]
\begin{small}
\caption{\textit{mini}-ImageNet and \textit{tiered}-ImageNet split statistics}
\vspace{-0.1in}
\label{tab:stats}
\begin{center}
\begin{tabular}{cc|crr|crr}
\toprule
&& \multicolumn{3}{c|}{\textit{mini}-ImageNet}& \multicolumn{3}{c}{\textit{tiered}-ImageNet} \\
Classes                & Purpose & Split         & N. Cls  & N. Img  & Split           & N. Cls   & N. Img \\
\midrule
\multirow{3}{*}{Base}  & Train   & Train-Train   & 64      & 38,400  & Train-A-Train   & 200      & 203,751   \\
                      & Val     & Train-Val     & 64      & 18,748  & Train-A-Val     & 200      & 25,460    \\
                      & Test    & Train-Test    & 64      & 19,200  & Train-A-Test    & 200      & 25,488    \\
\midrule
\multirow{3}{*}{Novel} & Train   & Train-Train   & 64      & 38,400  & Train-B         & 151      & 193,996   \\
                      & Val     & Val           & 16      & 9,600   & Val             & 97       & 124,261   \\
                      & Test    & Test          & 20      & 12,000  & Test            & 160      & 206,209   \\
\bottomrule
\end{tabular}
\end{center}
\end{small}
\vspace{-0.2in}
\end{table}

\subsection{Validation and testing splits for base classes}
In standard few-shot learning, meta-training, validation, and test set have disjoint sets of object
classes. However, in our incremental few-shot learning setting, to evaluate the model performance on
the base class predictions, additional splits of validation and test splits of the meta-training set
are required. Splits and dataset statistics are listed in Table~\ref{tab:stats}. For
\textit{mini}-ImageNet, \cite{lwof} released additional images for evaluating training set, namely
``Train-Val'' and ``Train-Test''. For \textit{tiered}-ImageNet, we split out $\approx$ 20\% of the
images for validation and testing of the base classes.

\subsection{Novel classes}
In \textit{mini}-ImageNet experiments, the same training set is used for both $\mathcal{D}_a$ and
$\mathcal{D}_b$. In order to pretend that the classes in the few-shot episode are novel, following
\cite{lwof}, we masked the base classes in $W_a$, which contains 64 base classes. In other words, we
essentially train for a 59+5 classification task. We found that under this setting, the progress
of meta-learning in the second stage is not very significant, since all classes have already been
seen before.

In \textit{tiered}-ImageNet experiments, to emulate the process of learning novel classes during the
second stage, we split the training classes into base classes (``Train-A'') with 200 classes and novel classes (``Train-B'') with 151 classes, just for meta-learning purpose.
During the first stage the classifier is trained using Train-A-Train data. In each meta-learning episode we sample few-shot examples from the novel classes (Train-B) and a query base set from Train-A-Val.  

{\bf 200 Base Classes (``Train-A''):}

{\tt n02128757, n02950826, n01694178, n01582220, n03075370, n01531178, n03947888, n03884397, n02883205, n03788195, n04141975, n02992529, n03954731, n03661043, n04606251, n03344393, n01847000, n03032252, n02128385, n04443257, n03394916, n01592084, n02398521, n01748264, n04355338, n02481823, n03146219, n02963159, n02123597, n01675722, n03637318, n04136333, n02002556, n02408429, n02415577, n02787622, n04008634, n02091831, n02488702, n04515003, n04370456, n02093256, n01693334, n02088466, n03495258, n02865351, n01688243, n02093428, n02410509, n02487347, n03249569, n03866082, n04479046, n02093754, n01687978, n04350905, n02488291, n02804610, n02094433, n03481172, n01689811, n04423845, n03476684, n04536866, n01751748, n02028035, n03770439, n04417672, n02988304, n03673027, n02492660, n03840681, n02011460, n03272010, n02089078, n03109150, n03424325, n02002724, n03857828, n02007558, n02096051, n01601694, n04273569, n02018207, n01756291, n04208210, n03447447, n02091467, n02089867, n02089973, n03777754, n04392985, n02125311, n02676566, n02092002, n02051845, n04153751, n02097209, n04376876, n02097298, n04371430, n03461385, n04540053, n04552348, n02097047, n02494079, n03457902, n02403003, n03781244, n02895154, n02422699, n04254680, n02672831, n02483362, n02690373, n02092339, n02879718, n02776631, n04141076, n03710721, n03658185, n01728920, n02009229, n03929855, n03721384, n03773504, n03649909, n04523525, n02088632, n04347754, n02058221, n02091635, n02094258, n01695060, n02486410, n03017168, n02910353, n03594734, n02095570, n03706229, n02791270, n02127052, n02009912, n03467068, n02094114, n03782006, n01558993, n03841143, n02825657, n03110669, n03877845, n02128925, n02091032, n03595614, n01735189, n04081281, n04328186, n03494278, n02841315, n03854065, n03498962, n04141327, n02951585, n02397096, n02123045, n02095889, n01532829, n02981792, n02097130, n04317175, n04311174, n03372029, n04229816, n02802426, n03980874, n02486261, n02006656, n02025239, n03967562, n03089624, n02129165, n01753488, n02124075, n02500267, n03544143, n02687172, n02391049, n02412080, n04118776, n03838899, n01580077, n04589890, n03188531, n03874599, n02843684, n02489166, n01855672, n04483307, n02096177, n02088364.}

{\bf 151 Novel Classes (``Train-B''):}

{\tt n03720891, n02090379, n03134739, n03584254, n02859443, n03617480, n01677366, n02490219, n02749479, n04044716, n03942813, n02692877, n01534433, n02708093, n03804744, n04162706, n04590129, n04356056, n01729322, n02091134, n03788365, n01739381, n02727426, n02396427, n03527444, n01682714, n03630383, n04591157, n02871525, n02096585, n02093991, n02013706, n04200800, n04090263, n02493793, n03529860, n02088238, n02992211, n03657121, n02492035, n03662601, n04127249, n03197337, n02056570, n04005630, n01537544, n02422106, n02130308, n03187595, n03028079, n02098413, n02098105, n02480855, n02437616, n02123159, n03803284, n02090622, n02012849, n01744401, n06785654, n04192698, n02027492, n02129604, n02090721, n02395406, n02794156, n01860187, n01740131, n02097658, n03220513, n04462240, n01737021, n04346328, n04487394, n03627232, n04023962, n03598930, n03000247, n04009552, n02123394, n01729977, n02037110, n01734418, n02417914, n02979186, n01530575, n03534580, n03447721, n04118538, n02951358, n01749939, n02033041, n04548280, n01755581, n03208938, n04154565, n02927161, n02484975, n03445777, n02840245, n02837789, n02437312, n04266014, n03347037, n04612504, n02497673, n03085013, n02098286, n03692522, n04147183, n01728572, n02483708, n04435653, n02480495, n01742172, n03452741, n03956157, n02667093, n04409515, n02096437, n01685808, n02799071, n02095314, n04325704, n02793495, n03891332, n02782093, n02018795, n03041632, n02097474, n03404251, n01560419, n02093647, n03196217, n03325584, n02493509, n04507155, n03970156, n02088094, n01692333, n01855032, n02017213, n02423022, n03095699, n04086273, n02096294, n03902125, n02892767, n02091244, n02093859, n02389026.}

\fi
\end{document}